\documentclass[journal]{IEEEtran}

\usepackage{times}
\usepackage{epsfig}
\usepackage{graphicx}
\usepackage{amsmath}
\usepackage{amssymb}
\usepackage{algorithmic}
\usepackage{multirow}
\usepackage[table,xcdraw]{xcolor}
\usepackage[ruled,vlined]{algorithm2e}

\usepackage[breaklinks=true,bookmarks=false]{hyperref}

\usepackage{caption}

\graphicspath{{Figs/}}

\newcommand{\etal}{\textit{et al.}}
\begin{document}
	
	\title{Semi-Supervised Image Deraining using Gaussian Processes}

	\author{Rajeev Yasarla,~\IEEEmembership{Student Member,~IEEE,}
		Vishwanath A. Sindagi,~\IEEEmembership{Student Member,~IEEE,}
		and~Vishal M. Patel,~\IEEEmembership{Senior~Member,~IEEE}
		\IEEEcompsocitemizethanks{\IEEEcompsocthanksitem R. Yasarla, V. A. Sindagi,  and V. M. Patel are with the Department of Electrical and Computer Engineering, Johns Hopkins University, Baltimore, MD, 21218.\protect\\
			E-mail: E-mail: ( ryasarla19@jhu.edu, vishwanathsindagi@jhu.edu, vpatel36@jhu.edu)}
		\thanks{Manuscript received...}}

	\maketitle

	\begin{abstract}
		 Recent CNN-based methods for image deraining  have achieved excellent performance in terms of reconstruction error as well as visual quality. However, these methods are limited in the sense that they can be trained only on fully labeled data. Due to various challenges in obtaining real world fully-labeled image deraining datasets, existing methods are trained only on synthetically generated data and hence, generalize poorly to real-world images. The use of real-world data in training image deraining networks is relatively less explored in the literature.  We propose a Gaussian Process-based semi-supervised learning framework which enables the network in learning to derain using synthetic dataset while generalizing better using  unlabeled real-world images. More specifically, we model the latent space vectors of unlabeled data using Gaussian Processes, which is then used to compute pseudo-ground-truth for supervising the network on unlabeled data. Through extensive experiments and ablations on several challenging datasets (such as Rain800, Rain200L and DDN-SIRR), we show that the proposed method is able to effectively leverage unlabeled data thereby resulting in significantly better performance as compared to labeled-only training. Additionally, we demonstrate that using unlabeled real-world images in the proposed GP-based framework results in superior performance as compared to the existing methods. 
	\end{abstract}
	

	\IEEEpeerreviewmaketitle
	\section{Introduction}
	Images captured under rainy conditions are often of poor quality. The artifacts introduced by rain streaks adversely affect the performance of subsequent computer vision algorithms such as object detection and recognition \cite{girshick2015fast,liu2016ssd, ren2015faster, Chen2018DomainAF}. With such algorithms becoming vital components in several applications such as autonomous navigation and video surveillance \cite{qi2018frustum,liang2018deep,perera2018uav}, it is increasingly important to develop algorithms for rain removal.

	\begin{figure}[t!]
		\begin{center}
			\includegraphics[width=.323\linewidth,height=0.242\linewidth]{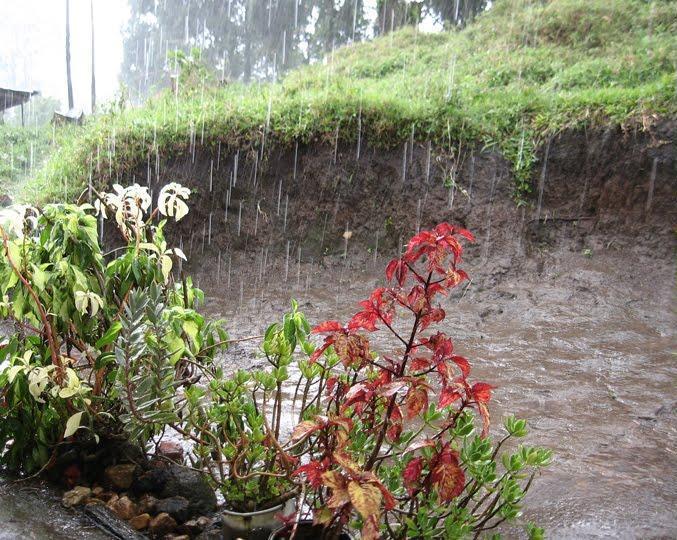}
			\includegraphics[width=.323\linewidth,height=0.242\linewidth]{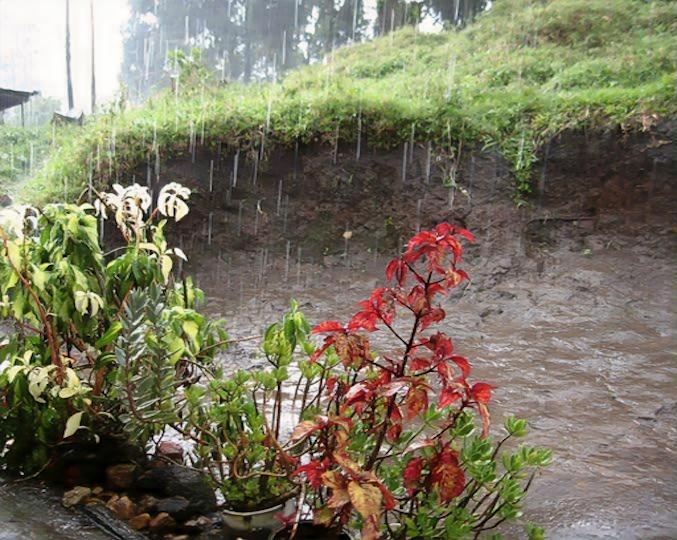}
			\includegraphics[width=.323\linewidth,height=0.242\linewidth]{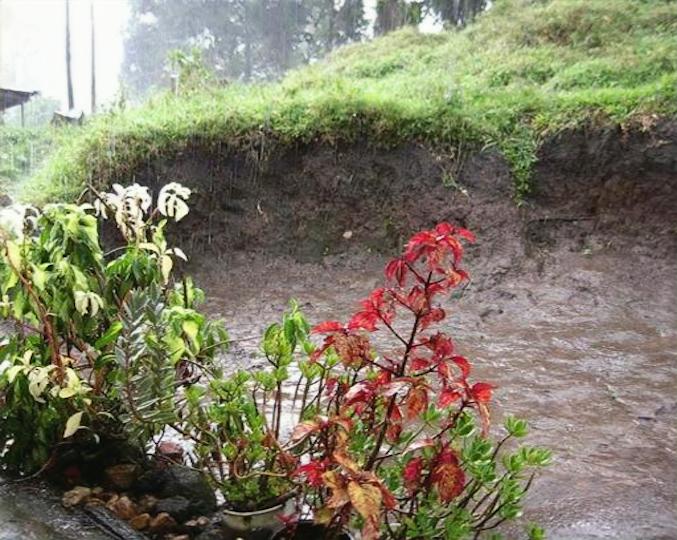}\\ \vskip2pt
			\includegraphics[width=.323\linewidth]{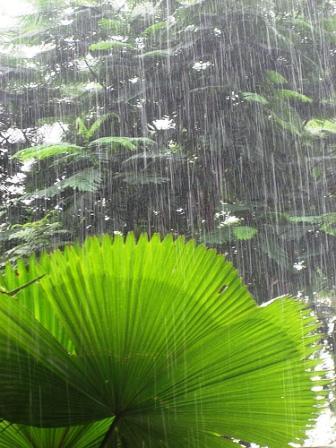}
			\includegraphics[width=.323\linewidth]{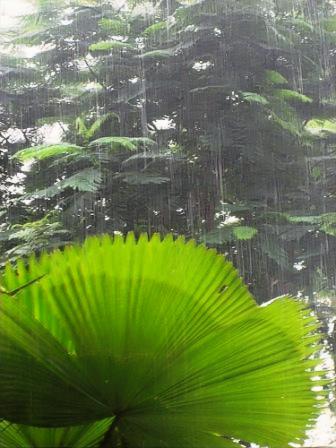}
			\includegraphics[width=.323\linewidth]{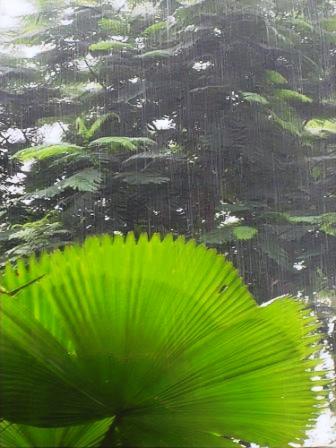}\\
			(a) Rainy Image \hskip30pt (b)  without GP \hskip30pt (c) With GP\\ 

		\end{center}
		\vskip -10pt \caption{ (a) Input rainy images. (b) Output from a network trained using only the synthetic data. (c)  Output from a network trained using the synthetic data and unlabeled real-world data. This shows better generalization.}
		\label{fig:img1}
	\end{figure}
	
	\begin{figure}[t!]
		\begin{center}
			\includegraphics[width=1\linewidth]{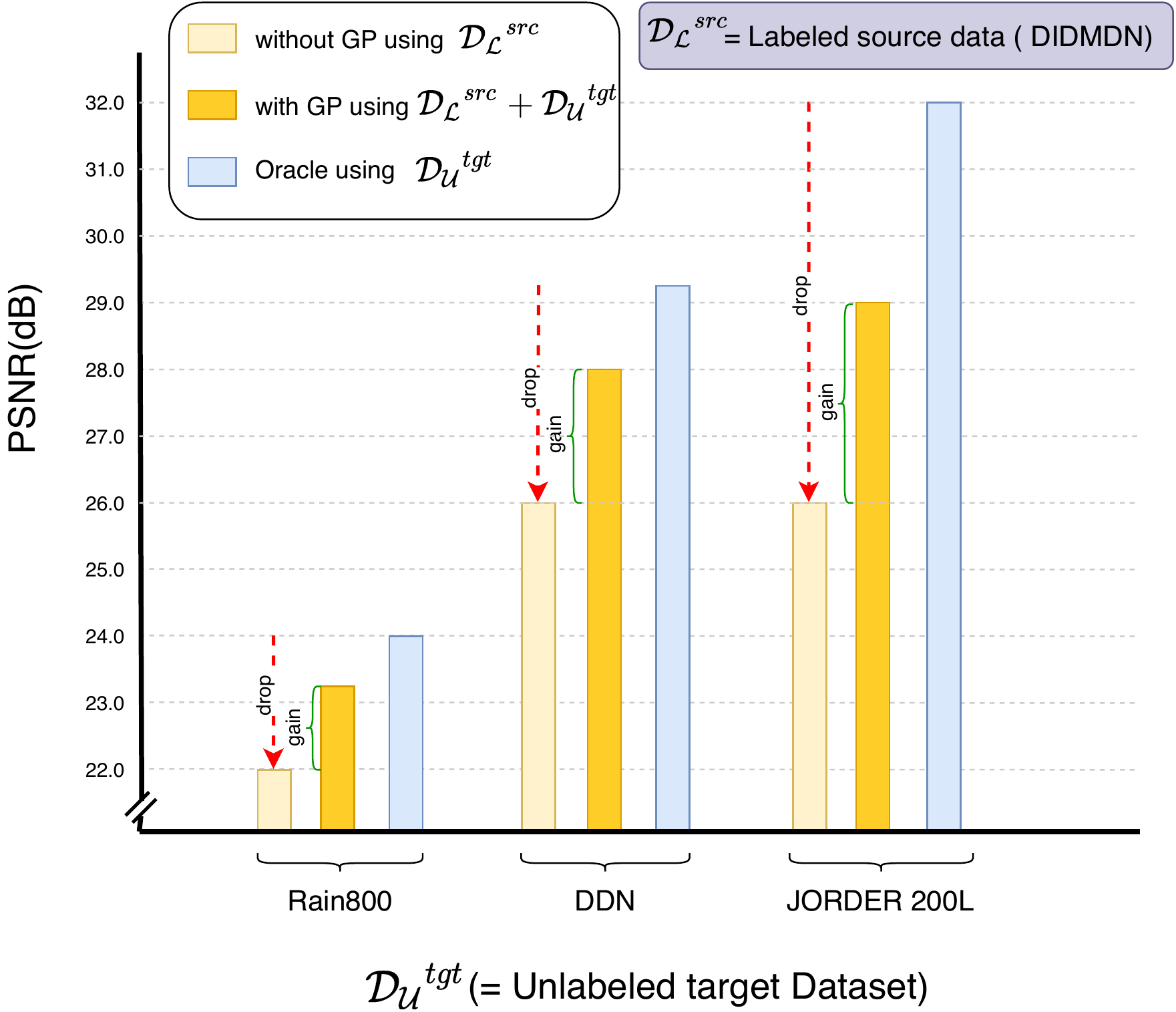} \vskip2pt
			
		\end{center}
		\vskip -10pt \caption{ Results from cross-domain semi-supervised learning (SSL)  experiments, where DIDMDN is used as labeled source data $\mathcal{D_{L}}^{src}$, and Rain800, DDN, and Rain200L as unlabeled target data $\mathcal{D_{U}}^{tgt}$ respectively. Oracle performance is when the base network trained in fully supervision fashion using unlabeled data.}
		\label{fig:img1_1}
	\end{figure}
	
	The task of rain removal is plagued with several issues such as (i) large variations in scale, density and orientation of the rain streaks, and (ii) lack of real-world labeled training data. Most of the existing work \cite{Authors18,Authors17f,Authors16,yang2017deep, Authors17c,fan2018residual,yang2019joint,li2019single} in image deraining have largely focused towards addressing the first issue. For example, Fu \etal \cite{Authors17f} developed an end-to-end method which focuses on high frequency detail during training a deraining network.  In another work, Zhang and Patel \cite{Authors18} proposed a density-aware multi-steam densely connected network for joint rain density estimation and deraining. Li \etal \cite{Authors18d} incorporated context information through recurrent neural networks for rain removal. More recently, Ren \etal \cite{ren2019progressive} introduced a progressive ResNet that leverages dependencies of features across stages.  While these methods have achieved superior performance in obtaining high-quality derained images, they are inherently limited due to the fact that they are fully-supervised networks and they can only leverage fully-labeled training data. However, as mentioned earlier, obtaining labeled real-world training data is quite challenging and hence, existing methods typically train their networks only on synthetically generated rain datasets \cite{Authors17e,yang2017deep}. 
	
	The use of synthetic datasets results in sub-optimal performance on the real-world images, typically because of the distributional-shift between synthetic and rainy images \cite{Chen2018DomainAF}.  Despite this gap in performance, this issue remains relatively unexplored in the literature.

	Recently, Wei \etal \cite{wei2019semi} proposed a semi-supervised learning framework (SIRR) where they simultaneously learn from labeled and unlabeled data for the purpose of image deraining. For training on the labeled data, they use the traditional mean absolute error loss between predictions and ground-truth (GT). For unlabeled data, they model the rain residual (difference between the input and output) through a likelihood term imposed on a Gaussian mixture model (GMM).  Furthermore, they enforce additional consistency that the distribution of synthetic rain is closer to that of real rain by minimizing the Kullback-Leibler (KL) divergence between them.  This is the first method to formulate the task of image deraining in a semi-supervised learning framework that can leverage unlabeled real-world images to improve the generalization capabilities. Although this method achieves promising results, it has the following drawbacks: (i) Due to the multi-modal nature of rain residuals, the authors assume that they  can be modeled using  GMM. This is true only if the actual residuals are being used to compute the GMM parameters. However, the authors use the  predicted rain residuals of real-world (unlabeled) images over training iterations for modeling the GMM. The  same model is then used to compute the likelihood of the predicted residuals (of unlabeled images) in the subsequent iterations. Hence, if the GMM parameters learned during the initial set of iterations are not accurate, which is most likely the case in the early stages of training, it will lead to sub-optimal performance. (ii) The goal  of using the KL divergence  is to bring the synthetic rain distribution  closer to the real rain distribution. As stated earlier, the predictions of real rain residuals will not be accurate during the earlier stages of training and hence, minimizing the discrepancy between the two distributions may not be appropriate. (iii) Using GMM to model the rain residuals requires one to choose the number of mixture components, rendering the model to be sensitive to such choices. 
	
\begin{figure}[htp!]
		\begin{center}
			\includegraphics[width=.323\linewidth,height=0.242\linewidth]{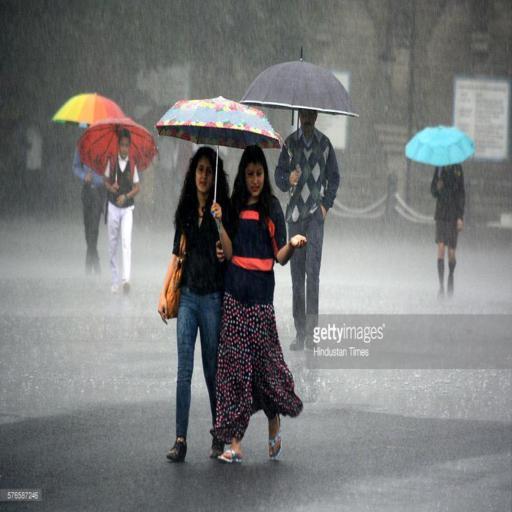}
			\includegraphics[width=.323\linewidth,height=0.242\linewidth]{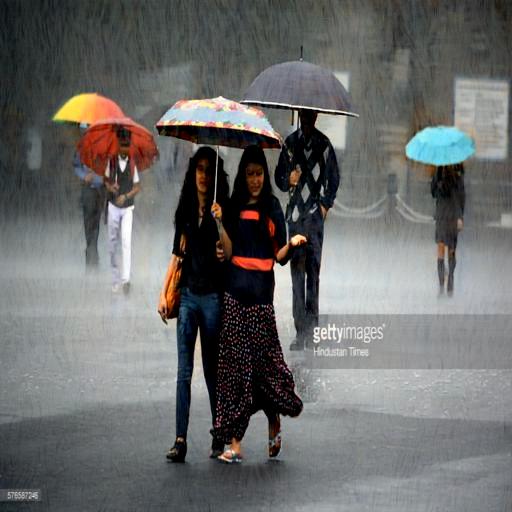}
			\includegraphics[width=.323\linewidth,height=0.242\linewidth]{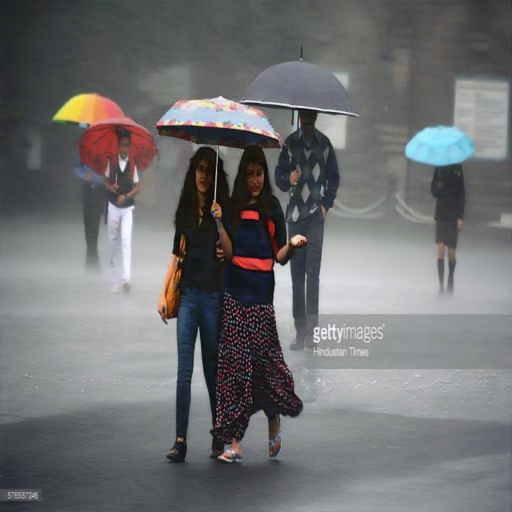}\\ \vskip2pt
			\includegraphics[width=.323\linewidth,height=0.242\linewidth]{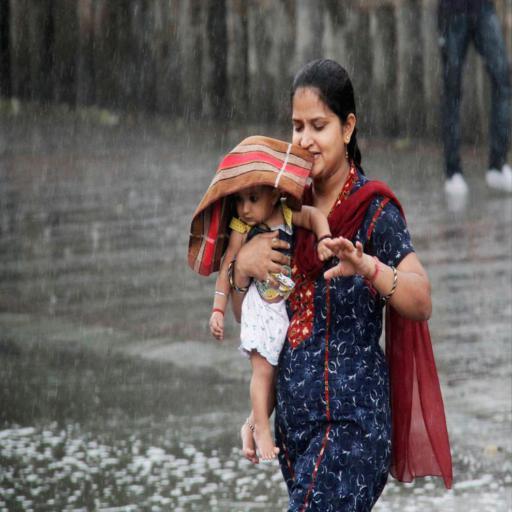}
			\includegraphics[width=.323\linewidth,height=0.242\linewidth]{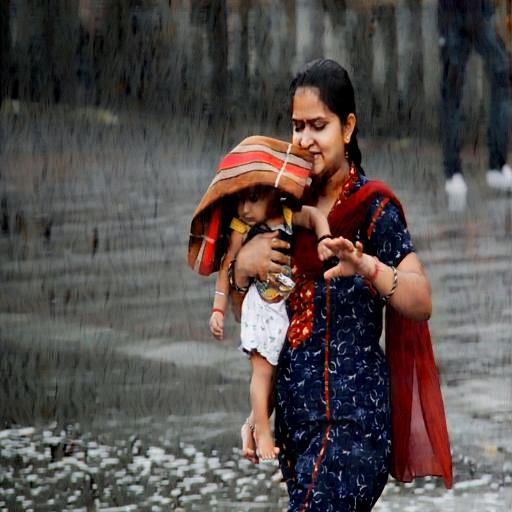}
			\includegraphics[width=.323\linewidth,height=0.242\linewidth]{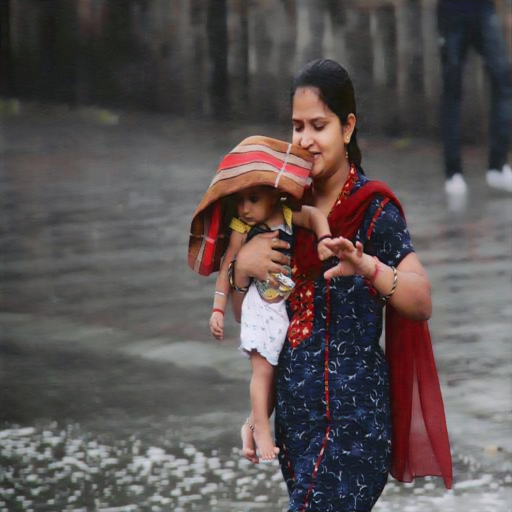}\\
			Rainy Image \hskip 55pt SIRR \hskip 55pt Ours
		\end{center}
		\caption{Sample deraining results using real-world rainy images. It can be observed that the proposed method achieves better deraining.}
		\label{fig:img2}
	\end{figure}

	Inspired by Wei \etal \cite{wei2019semi}, we address the issue of incorporating unlabeled real-world images into the training process for better generalization by overcoming the drawbacks of their method. In contrast to \cite{wei2019semi}, we use a non-parametric approach to generate supervision for the unlabeled data.  Specifically, we propose a Gaussian-process (GP) based semi-supervised learning (SSL) framework which involves iteratively training on the labeled and unlabeled data. The labeled learning phase involves training on the labeled data using  mean squared error between the predictions and the ground-truth. Additionally, inputs (from labeled dataset)  are projected onto the latent space, which are then modeled using GP. During the unlabeled training phase, we generate pseudo-GT for the unlabeled inputs using the GP modeled earlier in the labeled training phase. This pseudo GT is then used to supervise the intermediate latent space for the unlabeled data. The creation of the pseudo GT is based on the assumption that unlabeled images, when projected to the latent space, can be expressed as a weighted combination of the labeled data features where the weights are determined using a kernel function. These weights indicate the uncertainty of the labeled data points being used to formulate the unlabeled data point. Hence,  minimizing the error between the unlabeled data projections and the pseudo GT reduces the variance, hence resulting in the network weights being adapted automatically to the domain of  unlabeled data. Fig. \ref{fig:img1} and Fig.~\ref{fig:img1_1} demonstrates the results of leveraging unlabeled data using the proposed framework. Fig. \ref{fig:img2} compares the results of the proposed method with SIRR \cite{wei2019semi}. One can clearly see that our method is able to provide better results as compared to SIRR \cite{wei2019semi}. 
	To summarize, this paper makes the following contributions:
	\begin{itemize}
		\item We propose a non-parametric approach (which we call as Syn2Real) for performing SSL to  incorporate unlabeled real-world data into the training process. 
		\item The proposed method consists of modeling the intermediate latent space in the network using GP, which is then used to create the pseudo GT for the unlabeled data. The pseudo GT is further used to supervise the network at the intermediate level for the unlabeled data. 
		\item Through extensive experiments on different datasets,  we show that the proposed method is able to achieve on-par performance with limited training data as compared to network trained with full training data. Additionally, we also show that using the proposed GP-based SSL framework to incorporate the unlabeled real-world data into the training process results in better performance as compared to the existing methods. 
	\end{itemize}

		Note that this work is an extension of our previous work \cite{Yasarla_2020_CVPR} published in CVPR '20. In our earlier work,  we employed  single output GP for modeling the latent space vectors.  That is, the entire latent space vector (of dimension  $C\times W \times H$ ) of an unlabelled data point was formulated as a function (linear combination) of the latent space vectors of the labeled data. By doing so, different channels of the unlabeled latent vector are obtained using  the same set of weights, resulting in lower capacity of the function.  This leads to higher approximation error especially in cases where the individual feature maps of the unlabeled data vector are independent and hence require to be individually expressed as a function of the labeled data vectors. To address this issue, we model GP at feature level, where each  feature map of the latent vector of an unlabelled image is computed using labelled data latent vectors.  
		
		Additionally, we propose to  employ variance computed during the  GP formulation as a measure of uncertainty of the pseudo-GT, and use it further to weigh the loss function in order to ensure that only confident pseudo-GTs are used for training on the unlabeled data. To summarize, this paper makes the following improvements:
		\begin{itemize}
			\item Modeling Gaussian Process at feature level to account for  the fact that feature maps of unlabelled image can be independent. We name the new approach as Syn2Real++.
			\item Use of uncertainty to ensure only confident pseudo-GTs are used for supervising on the unlabeled data. 
			\item Additional experiments involving cross-dataset protocol to demonstrate that the proposed method is able to generalize well to cross-domain scenarios. 
			\item Additional ablation experiments involving  different kernels in   Gaussian Process.
		\end{itemize}
	
	\section{Related work}
	Image deraining is an extensively researched topic in the low-level computer vision community. Several approaches have been developed  to address this problem. These approaches are  classified into two main  categories: single image-based techniques \cite{Authors18,Authors17f,Authors16,yang2017deep, Authors17c,9007569}  and video-based techniques \cite{Authors6,Authors7,Authors15c,Authors18e,Authors18f,liu2018d3r}.  A comprehensive analysis of these methods can be found in \cite{li2019singlecomprehensive}.

	Single image-based techniques typically consume a single image as the input and attempt to reconstruct a rain-free image from it. Early methods for single image deraining either employed priors such as sparsity \cite{Authors17g, Authors15} and low-rank representation \cite{Low_Rank_ICCV2013} or modeled image patches using techniques such as dictionary learning \cite{Authors9} and GMM  \cite{Authors2000}. Recently, deep learning-based techniques have gained prominence due to their effectiveness in ability to learn efficiently from paired data. Video-based deraining techniques typically leverage additional information by enforcing  constraints like temporal consistency among the frames. 
	
	In this work, we focus on a semi-supervised single image-based deraining that specifically leverages additional unlabeled real-world data. Fu et al. \cite{Authors17d} proposed a convolutional neural network (CNN) based approach in which they learns a mapping  from a rainy image to  the clean  image.  Zhang et al. \cite{Authors17e} introduced  generative adversarial network (GAN) for image de-raining that resulted in high quality reconstructions.  Fu et al. \cite{Authors17f}  presented an end-to-end CNN called,    deep detail network, which directly reduces the mapping range from input to output.  Zhang and Patel \cite{Authors18}  proposed  a density-aware multi-stream densely connected CNN for joint rain density  estimation  and  deraining. Their network first classifies the input image based on the rain density, and then employs an appropriate network based on the predicted rain density to remove the rain streaks from the input image. Wang et al. \cite{Authors17b} employed  a hierarchical approach based on estimating different frequency details of an image to obtain the derained image.  Qian \etal \cite{Authors18b} proposed a GAN to remove rain drops from camera lens. To enable the network focus on important regions, they injected attention map into the generative and discriminating network.  Li et al. \cite{Authors18d} proposed a convolutional and recurrent neural network-based method for single image deraining that incorporates context information.    Recently, Li \etal \cite{li2019heavy}  and Hu \etal \cite{hu2019depth} incorporated depth information to improve the deraining quality. Yasarla and Patel \cite{yasarla2019uncertainty} employed uncertainty mechanism to learn location-based confidence for the predicted residuals. Wang \etal \cite{wang2019spatial} proposed a spatial attention network that removes rain in a local to global manner. Wang  \textit{et al.}\cite{wang2019erl} proposed a network that entangles the original low-quality embedding to a lantent vector which adaptively adds residues to obtain optimal latent vector in the residual learning branch. Kui~\cite{jiang2020multi} \etal proposed Multi-Scale Progressive Fusion Network where the multi-scale  representations for rain streaks from the perspective of input image scales and hierarchical deep features in a unified.

	\section{Background}
	
	In this section, we provide a formulation of the problem statement, followed by a brief description of key concepts in GP.
	
	\subsection{Single image de-raining} 
	Existing image deraining methods assume the additive model where the rainy image ($x$) is considered to be the superposition of a clean image ($y$) and a rain component ($r$), \textit{i.e},
	\setlength{\belowdisplayskip}{0pt} \setlength{\belowdisplayshortskip}{0pt}
	\setlength{\abovedisplayskip}{0pt} \setlength{\abovedisplayshortskip}{0pt}
	\begin{equation}
	x= y + r.
	\end{equation}
	Single image deraining task is typically an inverse problem where the goal is to   estimate the clean image $y$, given a rainy image $x$. This can be achieved by learning a function that either (i) directly maps from rainy image to clean image \cite{eigen2013restoring,fu2017clearing,Authors17c,Authors17g}, or (ii) extracts the rain component from the rainy image which can then be subtracted from the rainy image to obtain the clean image \cite{Authors17f,Authors18,li2018recurrent}. We follow the second approach of estimating the rain component from a rainy image.
	
	\subsection{Semi-supervised learning}
	In  semi-supervised learning, we are given a labeled dataset of input-target pairs ($\{x,y\} \in \mathcal{D_L}$) sampled from an unknown joint distribution $p(x,y)$  and unlabeled input data points $x \in \mathcal{D_U}$ sampled from $p(x)$. The goal is to learn a  function $f(x|\theta)$  
	parameterized by $\theta$ that accurately predicts the correct target $y$ for unseen samples from $p(x)$. The parameters $\theta$ are learned by leveraging both labeled and unlabeled datasets. Since the  labeled dataset  consists of input-target pairs, supervised loss functions such as mean absolute error or cross entropy are typically used to train the networks. The unlabeled datapoints form $\mathcal{D_U}$
	are used to augment $f(x|\theta)$   with information about the structure of $p(x)$ like shape of the data manifold \cite{oliver2018realistic} via different techniques such as enforcing consistent regularization \cite{laine2016temporal}, virtual adversarial training \cite{miyato2018virtual}  or pseudo-labeling \cite{lee2013pseudo}. 
	
	Following \cite{wei2019semi}, we employ the semi-supervised learning framework to leverage unlabeled real-world data to obtain better generalization performance. Specifically, we consider the synthetically generated rain dataset consisting of input-target pairs as the labeled dataset $\mathcal{D_L}$ and real-world unlabeled images as the unlabeled dataset $\mathcal{D_U}$. In contrast to \cite{wei2019semi}, we follow the approach of pseudo-labeling to leverage the unlabeled data.
	
	\subsection{Gaussian processes}
	A Gaussian process (GP) $f(v)$ is an infinite collection of random variables, of which any finite subset is jointly Gaussian distributed. A GP is completely specified by its mean function and covariance function which are defined as follows
	\begin{equation}
	\begin{aligned} m(v) &=\mathbb{E}[f(v)],
	\end{aligned}
	\end{equation}
	\begin{equation}
	\begin{aligned} {K}\left(v, v^{\prime}\right) &=\mathbb{E}\left[(f(v)-m(v))\left(f\left(v^{\prime}\right)-m\left(v^{\prime}\right)\right)\right], \end{aligned}
	\end{equation}
	where $v,v' \in \mathcal{V}$ denote the possible inputs that index the GP. The covariance matrix is constructed from a covariance function, or kernel, ${K}$ which expresses some prior notion of smoothness of the underlying function. GP can then be denoted as follows
	\begin{equation}
	f(v) \sim \mathcal{GP}(m(v), K(v, v')+\sigma_{\epsilon}^2I).
	\end{equation}
	where is I identity matrix and $\sigma_{\epsilon}^2$ is the variance of the additive noise.  Any collection of function values is then jointly Gaussian as follows
	\begin{equation}
	f(V)=\left[f\left(v_{1}\right), \ldots, f\left(v_{n}\right)\right]^{T} \sim \mathcal{N}\left(\mu, K(V, V')+\sigma_{\epsilon}^2I\right)
	\end{equation}
	with mean vector and covariance matrix defined by the GP as mentioned earlier. To make predictions at unlabeled points, one can compute a Gaussian posterior distribution in closed form by conditioning on the observed data. The reader is referred to \cite{rasmussen2003gaussian} for a  detailed review on GP.
	
			\begin{figure}[t!]
		\begin{center}
			\includegraphics[width=1\linewidth]{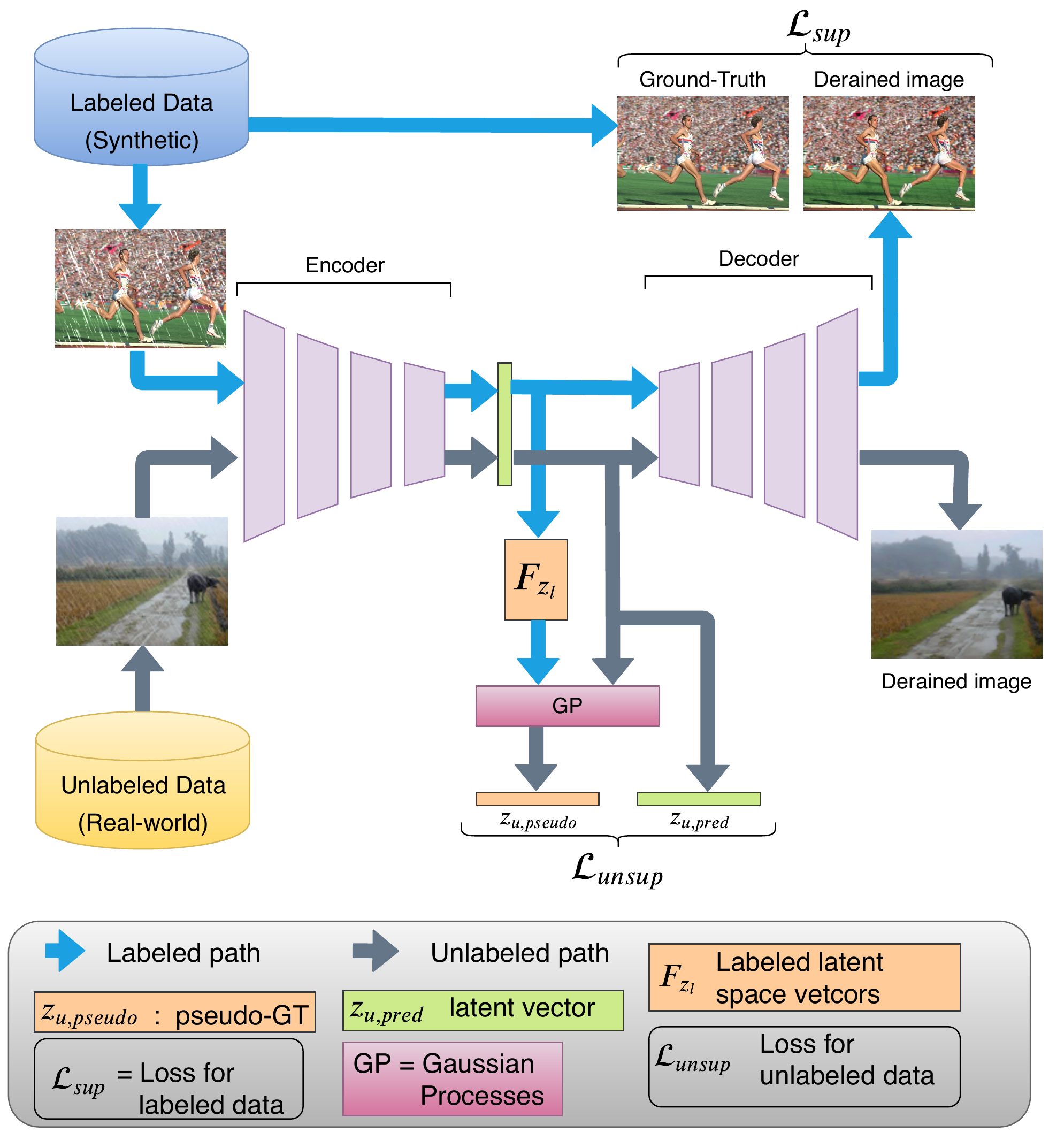} 
		\end{center}
		\vskip -10pt \caption{Overview of the proposed GP-based SSL framework. We leverage unlabeled data during learning. The training process consists of iterating over labeled data and unlabeled data. During the labeled training phase, we use supervised loss function consisting of $l_1$ error and perceptual loss between the prediction and targets. In the unlabeled phase, we jointly model the labeled and unlabeled latent vectors using GP to obtain the pseudo-GT for the unlabeled sample at the latent space. We use this pseudo-GT for supervision.}
		\label{fig:overview}
	\end{figure}

	\section{Proposed method}
	
	\begin{figure*}[t!]
		\begin{center}
			\centering
			\includegraphics[width=0.9\linewidth]{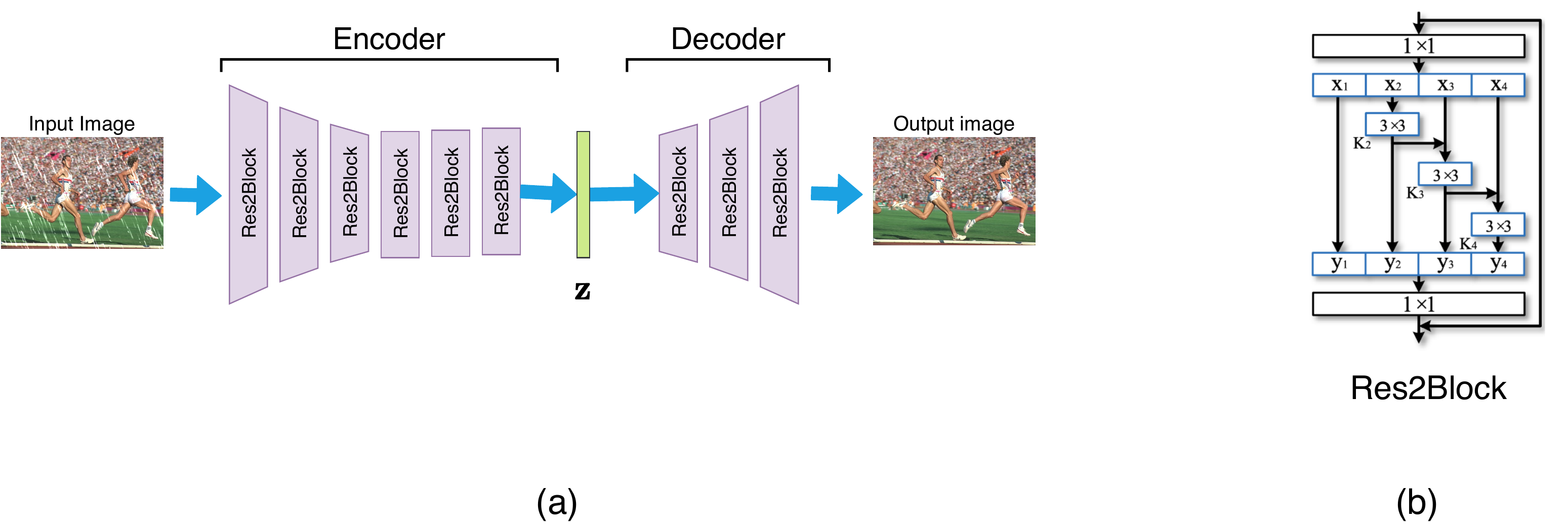} 
		\end{center}
		\vskip-15pt
		\caption{(a) U-Net based encoder and decoder with dense blocks (b) Res2Block~\cite{gao2019res2net} structure. Note that $\mathbf{z}$ is q latent vector.}
		\label{fig:denseblock}
	\end{figure*}


	The proposed method consists of a CNN based on the UNet structure \cite{ronneberger2015u}, where each block is constructed using a Res2Block~\cite{gao2019res2net} as shown in  Fig.~\ref{fig:denseblock}.
	The proposed network consists of an Encoder($h(.,\theta_{enc})$) and a Decoder($g(.,\theta_{dec})$). The encoder network ($h(.,\theta_{enc})$) consists of the following components:\\
	Conv2d $3\times 3$(3,16) - Res2Block(16,32) - Downsample - Res2Block(32,32) - Downsample - Res2Block(32,32) - Downsample - Res2Block(32,64) - Res2Block(64,64) - Res2Block(64,64).\\
	The decoder network ($g(.,\theta_{dec})$) of the following components:\\
	Res2Block(64,32) - Upsample - Res2Block(32,32) - Upsample - Res2Block(32,16) - Upsample - Conv2d $3\times 3$(16,3),\\
	where Conv2d $3\times 3(m, n)$ is a $3\times 3$ convolutional layer with $m$ input channels and $n$ output channels,Res2Block$(m, n)$ means Res2Block with $m$ input channels and $n$ output channels. In summary, the network is  made up of an encoder ($h(x,\theta_{enc})$) and a decoder ($g(\mathbf{z},\theta_{dec})$). The encoder and decoder networks are parameterized by $\theta_{enc}$ and $\theta_{dec}$, respectively. 
	Furthermore, $x$ is the input to the network  which is then mapped by the encoder to a latent feature matrix $\mathbf{z}$, \textit{i.e} each latent feature map ($z$) is vectorized arranged as rows in $\mathbf{z}=[z_1^T,z_2^T,...]^T$. In our case, $x$ is the rainy image from which we want to remove the rain streaks. The latent feature matrix is then fed to the decoder to produce the output $r$, which in our case is the rain streaks. The rain streak component is then subtracted form the rainy image ($x$) to produce the clean image ($y$), \textit{i.e},
	\begin{equation}
	y = x - r,
	\end{equation}
	where
	\begin{equation} 
	r = g(h(x,\theta_{enc}), \theta_{dec}).
	\end{equation}

	In our problem formulation, the training dataset is $\mathcal{D}=\mathcal{D_L} \cup \mathcal{D_U}$, where $\mathcal{D_L}=\{x_l^i,y_l^i\}_{i=1}^{N_l}$ is a labeled training set consisting of $N_{l}$ samples and $\mathcal{D_U}=\{x_u^i\}_{i=1}^{N_u}$ is a set consisting of $N_{u}$ unlabeled samples.  For the rest of the paper, $\mathcal{D_L}$ refers to   labeled ``\textit{synthetic}'' dataset and  $\mathcal{D_U}$ refers  to unlabeled ``\textit{real-world}'' dataset, unless otherwise specified. 
	
	The goal of the proposed method is to learn the network parameters by leveraging both labeled  ($\mathcal{D_L}$) and unlabeled dataset ($\mathcal{D_U}$). The training process iterates over labeled and unlabeled datasets. The network parameters are learned by minimizing (i) the supervised loss function ($\mathcal{L}_{sup}$) in the labeled training phase, and (ii) the unsupervised loss function ($\mathcal{L}_{unsup}$) in the unlabeled training phase.  For the unlabeled training phase, we generate pseudo GT using GP formulation, which is then used in the unsupervised loss function. The two training phases are described in detail in the following sections. 
	
	\subsection{Labeled training phase}
	In this phase, we use the labeled data $\mathcal{D_L}$ to  learn the network parameters. Specifically, we  minimize the following supervised loss function 
	\begin{equation}
	\label{eq:loss_sup}
	\mathcal{L}_{sup} = \mathcal{L}_1 + \lambda_p \mathcal{L}_{p},
	\end{equation}
	where $\lambda_{p}$ is a constant, and $\mathcal{L}_1$ and $\mathcal{L}_p$ are $l_1$-loss and perceptual loss~\cite{johnson2016perceptual,zhang2018multi} functions, respectively. They are defined as follows
	\begin{equation}
	\mathcal{L}_{1} = \|y^{pred}_l - y_l\|_1,
	\end{equation}
	\begin{equation}
	\mathcal{L}_{p} = \|\Phi_{VGG}(y^{pred}_l) - \Phi_{VGG}(y_l)\|^2_2 ,
	\end{equation}
	where $y^{pred}_l=g(z,\theta_{dec})$ is the predicted output,  $y_l$ is the ground-truth,  $\mathbf{z}=h(x, \theta_{enc})$ is the intermediate latent space vector and $\Phi_{VGG}(\cdot)$ represents the pre-trained VGG-16 \cite{simonyan2014very} network. For more details on the perceptual loss, please refer to supplementary material. 
	
	In addition to minimizing the loss function, we also store the intermediate feature vectors $\mathbf{z}_l^i$'s for all the labeled training images $x_l^i$'s in a matrix $\mathbf{F}_{\mathbf{z}_l}$. That is $\mathbf{F}_{\mathbf{z}_l}= \{z_l^i\}_{i=1}^{N_l}$. It is used later in the unlabeled training phase to generate the pseudo-GT for the unlabeled data. In our case,  $\mathbf{z}_l^i$ is a matrix of size $32\times 1024$, for the network in our proposed method. Thus $\mathbf{F}_{\mathbf{z}_l}$ is a matrix of size  $(32\times N_l)\times 1024$.
	
	\subsection{Unlabeled training phase}

\begin{figure}[htp!]
			\begin{center}
				\includegraphics[width=.44\linewidth]{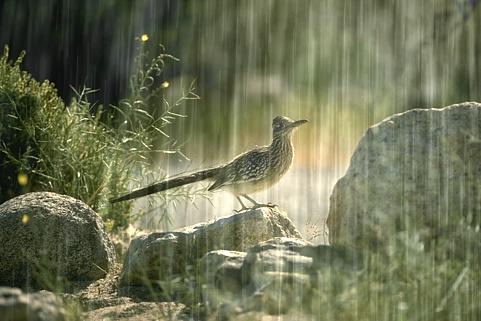}
				\includegraphics[width=.44\linewidth]{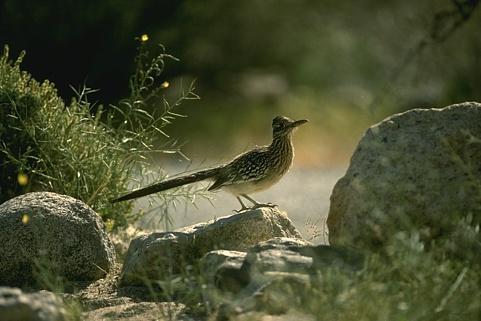}\\ 
				Rainy Image \hskip 55pt Ground-truth 
				\vskip2pt
				\includegraphics[width=.44\linewidth]{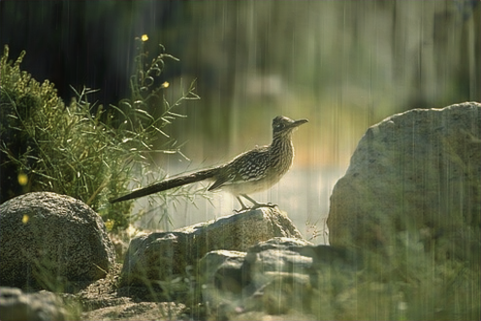}
				\includegraphics[width=.44\linewidth]{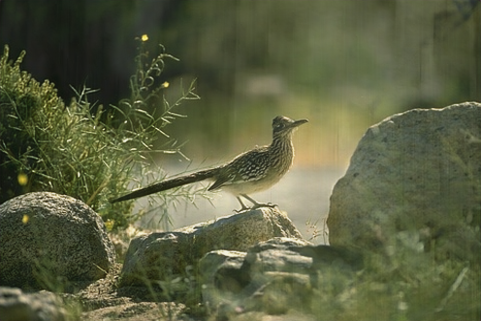}\\
				Syn2Real~\cite{Yasarla_2020_CVPR} \hskip 65pt Ours
			\end{center}
			\caption{Derained results using real-world rainy images. Our output. It can be observed that the proposed method achieves better deraining.}
			\label{fig:comp_gps}
		\end{figure}

		In this phase, we leverage the unlabeled data $\mathcal{D_U}$ to improve the generalization performance. Specifically, we provide supervision at the intermediate latent space feature matrix $\mathbf{z}_u$ by minimizing the error between the predicted latent feature $\mathbf{z}_u^{pred}$ and the pseudo-GT $\mathbf{z}_u^{pseudo}$  obtained by modeling the latent space vectors of the labeled sample images $\mathbf{F}_{\mathbf{z}_l}$ and $\mathbf{z}^{pred}_u$ jointly using GP.  In our previous work \cite{Yasarla_2020_CVPR}, we expressed $\mathbf{z}_u$ as a weighted combination of $\mathbf{z}_l$ and modelled joint distribution considering $\mathbf{z}_u$, and $\mathbf{z}_l$. Due to this, every feature map $z_{u1},z_{u2},...,z_{uM}$ in $\mathbf{z}_u$ will have the same weighted combination of $\mathbf{z}_l$. In other words, \cite{Yasarla_2020_CVPR} does not account for the fact that feature maps ($z_{u1},z_{u2},...,z_{uM}$) in the latent space feature matrix ($\mathbf{z}_u$) can be potentially independent.  This might result under or over deraining in some cases as shown in  Fig.~\ref{fig:comp_gps}. We overcome this issue by  expressing each feature map $z_{u1},z_{u2},...,z_{uM}$ as a different function of $\mathbf{z}_l$ like in Eq.~\eqref{eq:featuremap_rel}. Thus we obtain pseudo-GT ($z_{u1}^{pseudo},z_{u2}^{pseudo},...,z_{uM}^{pseudo}$ in $\mathbf{z}_u^{pseudo}$) for every feature map $z_{u1},z_{u2},...,z_{uM}$ in $\mathbf{z}_u$.
		
		\noindent\textbf{Pseudo-GT using GP:} The training occurs in an iterative manner, where we first learn the weights using   the labeled data ($\mathcal{D_L}$) followed by weight updates using the unlabeled data ($\mathcal{D_U}$). After the first iteration on  $\mathcal{D_L}$, we store the latent space matrices of the labeled data  in a list $\mathbf{F}_{\mathbf{z}_l}$. The latent space feature matrix $\mathbf{z}$ lies on a low dimension manifold. During the unlabeled phase, we project each feature map in the latent space feature matrix ($\mathbf{z}_u$) of the unlabeled input onto the space of labeled vectors $\mathbf{F}_{\mathbf{z}_l}= \{\mathbf{z}_l^i\}_{i=1}^{N_l}$. In our previous work \cite{Yasarla_2020_CVPR} we expressed $\mathbf{z}_u$ as weighted combination of $\mathbf{z}_l$ as follows:
		\begin{equation}
		\label{eq:lin_combination}
		z^k_u = \sum_{i=1}^{N_l} \alpha_i z_l^i +\epsilon,
		\end{equation}
		where $\alpha_{i}$ are the coefficients, and $\epsilon$ is additive noise $\mathcal{N}(0,\sigma_\epsilon^2)$.  
		
		However, every latent space feature vector in $\mathbf{z}_u$ may not have the same relation with every other labelled latent feature map in $\mathbf{z}_l$. To incorporate this fact, we express each feature map  $z_{u1}^k,z_{u2}^k,...,z_{uM}^k$ in $\mathbf{z_u}^k$ as follows,
		\begin{equation}
		\label{eq:featuremap_rel}
		\begin{aligned}
		z_{u,1}^k &= f_1(\mathbf{z}_l)+\epsilon,\\
		z_{u,2}^k &= f_2(\mathbf{z}_l)+\epsilon\\
		&\dots\\
		z_{u,M}^k &= f_M(\mathbf{z}_l)+\epsilon,\\
		\end{aligned}
		\end{equation}
		where $M$ is the number of feature map vectors in $\mathbf{z}$, and $f_1, f_2,\dots,f_M$ are different from each other. With this formulation, we can jointly model the distribution of the latent space feature matrices of the labeled and the unlabeled samples using GP.
	
	\begin{equation}
	\left[\begin{array}{l}
	\mathbf{z}_{l}\\
	\mathbf{z}_{u}
	\end{array}\right]=\mathcal{N}\left(\left[\begin{array}{l}
	\boldsymbol{\mu}_{l} \\
	\boldsymbol{\mu}_{u}
	\end{array}\right],\left[\begin{array}{cc}
	K\left(\mathbf{z}_{l}, \mathbf{z}_{l}\right) & K\left(\mathbf{z}_{l}, \mathbf{z}_{u}\right) \\
	K\left(\mathbf{z}_{u},\mathbf{z}_{l}\right) & K\left(\mathbf{z}_{u},\mathbf{z}_{u}\right)
	\end{array}\right]\right).
	\end{equation}\\
	$K(.,.)$  is the kernel matrix function, defined as follows 
	\begin{equation}
	K(\mathbf{z},\mathbf{z})_{k,i}= \kappa(z_k,z_i) = \frac{ \langle z_k, z_i\rangle}{|z_k|\cdot|z_i|}.
	\end{equation}
	Conditioning the joint distribution will yield the following conditional multi-variate Gaussian distribution for the unlabeled sample
	\begin{equation}
	P(\mathbf{z}^k_{u}|\mathcal{D_L},\mathbf{F}_{\mathbf{z}_l}) = \mathcal{N}(\boldsymbol{\mu}_u^k,\boldsymbol{\Sigma}_u^k),
	\end{equation}
	where 
	\begin{equation}
	\label{eq:mean}
	\boldsymbol{\mu}_u^k = K(\mathbf{z}_{u}^k, \mathbf{F}_{\mathbf{z}_l}) [K(\mathbf{F}_{\mathbf{z}_l},\mathbf{F}_{\mathbf{z}_l}) + \sigma_\epsilon^2 \mathbf{I}]^{-1}\mathbf{F}_{\mathbf{z}_l},
	\end{equation}
	\vskip2pt
	\begin{equation}
	\label{eq:sigma}
	\begin{aligned}
	\boldsymbol{\Sigma}_u^k = {} & K(\mathbf{z}_{u}^k,\mathbf{z}_{u}^k) - K(\mathbf{z}_{u}^k,\mathbf{F}_{\mathbf{z}_l})[K(\mathbf{F}_{\mathbf{z}_l},\mathbf{F}_{\mathbf{z}_l})+\sigma_\epsilon^2\mathbf{I}]^{-1} \\
	& K(\mathbf{F}_{\mathbf{z}_l},\mathbf{z}_{u}^k) + \sigma_\epsilon^2\mathbf{I},
	\end{aligned}
	\end{equation}
	where $\sigma_\epsilon^2$ is set equal to 1, 
	
	Note that $\mathbf{F}_{\mathbf{z}_l}$ contains the latent space vectors of all the labeled images, $K(\mathbf{F}_{\mathbf{z}_l},\mathbf{F}_{\mathbf{z}_l})$ is a matrix of size $(32\times N_l)\times (32\times N_l)$, and $K(\mathbf{z}_{u}^k,\mathbf{F}_{\mathbf{z}_l})$ is a matrix of size $32\times (32\times N_l)$. Using all the vectors may not be necessarily optimal for the following reasons: (i) These vectors will correspond to different regions in the image with a wide diversity in terms of content and density/orientation of rain streaks. It is important to consider only those vectors that are similar to the unlabeled vector.  (ii) Using all the vectors is computationally prohibitive. Hence, we use only $N_n$ nearest labeled vectors corresponding to an unlabeled vector. More specifically, we replace  $\mathbf{F}_{\mathbf{z}_l}$ by $\mathbf{F}_{\mathbf{z}_l,n}$ in Eq. \eqref{eq:featuremap_rel}-\eqref{eq:sigma}. Here $\mathbf{F}_{\mathbf{z}_l,n} =\{\mathbf{z}_l^j : \mathbf{z}_l^j \in nearest(\mathbf{z}_u^k,\mathbf{F}_{\mathbf{z}_l} ,N_n) \}$ with $nearest(p,Q ,N_n)$ being a function that finds top $N_n$ nearest neighbors of $p$ in $Q$.  
	
	\begin{table*}[ht!]
		\caption{Effect of using unlabeled real-world data in training process on DDN-SIRR dataset. Evaluation is performed on synthetic dataset similar to \cite{wei2019semi}. Proposed method achieves better gain in PSNR as compared to SIRR\cite{wei2019semi} in the case of both Dense and Sparse categories. SSL indicates semi-supervised learning. Gains are indicated in the brackets for SSL based methods.}
		\label{tab:ddnsirr_synthetic}
		\centering
		\resizebox{1\linewidth}{!}{
			\begin{tabular}{|l|c|ccccccc|cc|ccc|}
				\hline
				\multirow{3}{*}{{Dataset}} & \multirow{3}{*}{\begin{tabular}[c]{@{}c@{}}Input\end{tabular}} & \multicolumn{7}{c|}{Methods that use only synthetic dataset}                                                                                  & \multicolumn{5}{c|}{Methods that use synthetic and real-world dataset} \\ \cline{3-14} 
				&                                                                         & \multirow{2}{*}{\begin{tabular}[c]{@{}c@{}}DSC \cite{luo2015removing}\\(ICCV '15)\end{tabular}} & \multirow{2}{*}{\begin{tabular}[c]{@{}c@{}}LP \cite{li2016rain}\\(CVPR '16)\end{tabular}} & \multirow{2}{*}{\begin{tabular}[c]{@{}c@{}}JORDER \cite{yang2017deep}\\(CVPR '17)\end{tabular}} & \multirow{2}{*}{\begin{tabular}[c]{@{}c@{}}DDN \cite{Authors17f}\\(CVPR '17)\end{tabular}} & \multirow{2}{*}{\begin{tabular}[c]{@{}c@{}}JBO \cite{Authors17c}\\(CVPR '17)\end{tabular}} & \multirow{2}{*}{\begin{tabular}[c]{@{}c@{}}DID-MDN \cite{Authors18}\\(CVPR '18)\end{tabular}} & \multirow{2}{*}{\begin{tabular}[c]{@{}c@{}}UMRL \cite{yasarla2019uncertainty}\\(CVPR '19)\end{tabular}} & \multicolumn{2}{c|}{SIRR \cite{wei2019semi} (CVPR '19)}  & \multicolumn{3}{c|}{Ours}                 \\ \cline{10-14} 
				&                                                                         &                      &                     &                         &                      &                      &                          & & w/o SSL     & w/ SSL   & w/o SSL    & Syn2Real\cite{Yasarla_2020_CVPR}       & Syn2Real++            \\ \hline
				Dense                    & 17.95                                                                   & 19.00                & 19.27               & 18.75                   & 19.90                & 18.87                & 18.60   & 20.11                 & 20.01   & 21.60(1.59)     & 20.24  & 22.36(0.73) & \textbf{22.49(1.25)}    \\
				Sparse                   & 24.14                                                                   & 25.05                & 25.67               & 24.22                   & 26.88                & 25.24                & 25.66  &  26.94      & 26.90   & 26.98(0.08)    & 26.15  & 27.12(0.97) & \textbf{27.38(1.23)}  \\ \hline
			\end{tabular}
		}
	\end{table*}

		We use the mean predicted by Eq. \eqref{eq:mean} as the pseudo-GT ($\mathbf{z}_{u,pseudo}^{k}$) for supervision at the latent space level. By minimizing the error between $\mathbf{z}^{k}_{u,pred}=h(x_u,\theta_{enc})$ and  $\mathbf{z}_{u,pseudo}^{k}$, we update the weights of the encoder $h(\cdot,\theta_{enc})$, thereby adapting the network to unlabeled data which results in better generalization. Erroneous pseudo-GT ($\mathbf{z}_{u,pseudo}^{k}$)  computed using Gaussian Processes might limit the performance of the network. To address this, we re-weight L2 error between $\mathbf{z}^{k}_{u,pred}=h(x_u,\theta_{enc})$ and  $\mathbf{z}_{u,pseudo}^{k}$ using the predicted variance in \eqref{eq:sigma}. Additionally, we minimize the prediction variance by  including  Eq. \eqref{eq:sigma} in the overall loss function. Using GP we are approximating $\mathbf{z}_{u}^{k}$, latent vector of an unlabeled image using the latent space vectors in $\mathbf{F}_{\mathbf{z}_l}$, by doing this we may end up computing incorrect pseudo-GT predictions because of the dissimilarity between the latent vectors. This dissimilarity is due to different compositions in rain streaks like different densities, shapes, and directions of rain streaks. In order to address this issue, we minimize the variance $\boldsymbol{\Sigma}_{u,n}^k$ computed between $\mathbf{z}^{k}_{u}$ and the $N_n$ nearest neighbors in the latent space vectors using GP. 
	
	Thus, the loss used during training using the unlabeled data is defined as follows
	\begin{equation}
	\begin{aligned}
	\mathcal{L}_{unsup} =& (\mathbf{z}^{k}_{u,pred} - \mathbf{z}_{u,pseudo}^{k})^T \left(\boldsymbol{\Sigma}_u^k\right)^{-1} (\mathbf{z}^{k}_{u,pred} - \mathbf{z}_{u,pseudo}^{k}) \\
	&+ \log |\boldsymbol{\Sigma}_u^k|,
	\end{aligned}
	\end{equation}
	where $\mathbf{z}^{k}_{u,pred}$ is the latent vector obtained by forwarding an unlabeled input image $x_u^k$ through the encoder $h$, \text{i.e}, $\mathbf{z}^{k}_{u,pred}=h(x_u,\theta_{enc})$ , $\mathbf{z}_{u,pseudo}^{k} = \boldsymbol{\mu}_u^k$ is the  pseudo-GT latent space vector (see Eq. \eqref{eq:mean}), and $\boldsymbol{\Sigma}_{u,n}^{k}$ is the variance obtained by replacing $\mathbf{F}_{\mathbf{z}_l} $ in Eq. \eqref{eq:sigma} with $\mathbf{F}_{\mathbf{z}_l,n} $.
	
	\subsection{Total loss}
	The overall loss function used for training the network is defined as follows
	\begin{equation}
	\label{eq:loss_total}
	\mathcal{L}_{total} = \mathcal{L}_{sup} + \lambda_{unsup} \mathcal{L}_{unsup},
	\end{equation} 
	where $\lambda_{unsup}$ is a pre-defined weight that controls the contribution from $\mathcal{L}_{sup}$ and $\mathcal{L}_{unsup}$.
	
	\subsection{Training and implementation details}
	We use the UDeNet network that is  based on the UNet style encoder-decoder architecture \cite{ronneberger2015u} with a slight difference in the building blocks. The network  is trained using the Adam optimizer with a learning rate of 0.0002 and batch size of 4 for a total of 60 epochs. Furthermore, we reduce the  learning rate by a factor of 0.5 at every 30 epochs. We use $\lambda_p=0.04$ (Eq. \eqref{eq:loss_sup}), $\lambda_{unsup}=1.5 \times 10^{-4}$ (Eq. \eqref{eq:loss_total}), $N_n=64$. During training, the images are randomly cropped to the size of 256$\times$256. Ablation studies with different hyper-parameter values are provided in section~\ref{experiments}.

	\section{Experiments and results}\label{experiments}
	In this section, we present the details of the datasets and various experiments conducted to demonstrate the effectiveness of the proposed framework.  Specifically, we conducted two sets of experiments. In the first set, we analyze the effectiveness of using the unlabeled real-world data during training using the proposed framework. Here, we compare the performance of our method with a recent SSL framework for image deraining (SIRR) \cite{wei2019semi}. In the second set of experiments, we evaluate the proposed method by training it in cross domain fashion \textit{i.e} DIDMDN dataset is used as labeled images, and other datasets like Rain800, Rain200L, and DDN as unlabeled images. These experiments shows that our proposed method is effective in transferring knowledge from one rain distribution to another rain distribution.

		\subsection{Datasets}
		\noindent\textbf{DIDMDN:} The synthetic rain dataset DIDMDN published by the authors of \cite{Authors18} contains a total of 13,200 images with different densities of rain, \textit{i.e.} low, medium and high densities. These images are split into 12,000 training images and 1,200 test images. The test split contains 4,00 images in each of the low, medium and high density categories.\\ 
		
		\noindent\textbf{DDN dataset:}  The DDN dataset  consists of 9,100 image pairs obtained by synthesizing  different types of rain streaks on the clean images from the UCID dataset \cite{schaefer2003ucid}. Test set contains 1,400 images with different rain streak directions and densities\\
		
		\noindent\textbf{Rain800:} This dataset was published by Zhang \etal \cite{Authors17e} which contains 8,00 images in total. These 8,00 images are split into train and test splits. Train split consists of 7,00 paired rainy and corresponding real-world clean images.  The test set contains 1,00 rainy and corresponding clean images. \\
		
		\noindent\textbf{Rain200L:} Yang et al. \cite{yang2017deep} collected images from BSD200 \cite{martin2001database} to create 3 datasets:  Rain12, Rain200L and Rain200H. We use Rain200L for our cross domain experiments. Rain200L consists of 1,800 training images and 2,00 testing images.\\
		
		\noindent\textbf{DDN-SIRR dataset:} Wei \etal \cite{wei2019semi} constructed a dataset consisting of labeled synthetic training set and unlabeled real-world dataset. This dataset is constructed specifically to evaluate semi-supervised learning frameworks. DDN dataset is used as labelled training data, and  the unlabeled real-world synthetic train set comprises of images collected from \cite{wei2017should,yang2017deep,Authors17e} and Google image search. Furthermore, the test set consists of two categories: (i) Dense rain streaks, and (ii) Sparse rain streaks. Each test set  consists of 10 images.

	\subsection{Use of real-world data}
	The goal of this experiment is to analyze the effect of using unlabeled real-world data along with labeled synthetic dataset in the training framework. Following the protocol set by \cite{wei2019semi}, we use the ``labeled synthetic" train set from the DDN-SIRR dataset as  $\mathcal{D_L}$ and the ``real-world" train set from the DDN-SIRR dataset as $\mathcal{D_U}$. Evaluation is performed on (i) Synthetic test set from DDN-SIRR,  and (ii) Real-world test set from DDN-SIRR.\\

	\begin{figure*}[t!]
		\begin{center}
			\begin{minipage}[c]{0.24\linewidth}
				\begin{center}
					\includegraphics[width=\linewidth]{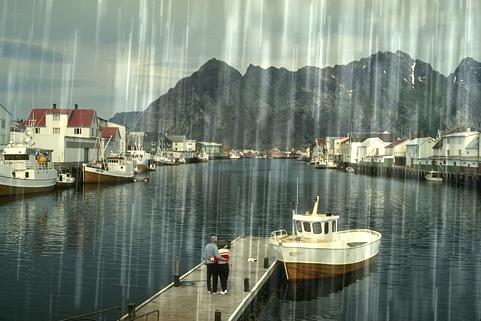}\\
					{\large Rainy Image}
				\end{center}
			\end{minipage}
			\hfill
			\begin{minipage}[c]{0.74\linewidth}
				\begin{center}
					{\large DIDMDN\cite{Authors18} \hskip65pt DDN\cite{Authors17f} \hskip70pt SIRR\cite{wei2019semi} }\\
					{\large (CVPR'18) \hskip65pt (CVPR'17) \hskip70pt (CVPR'19) }\\
					\includegraphics[width=.32\linewidth]{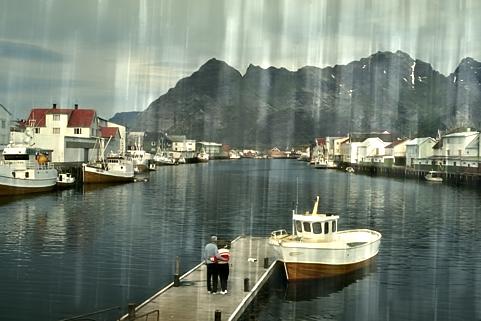}
					\includegraphics[width=.32\linewidth]{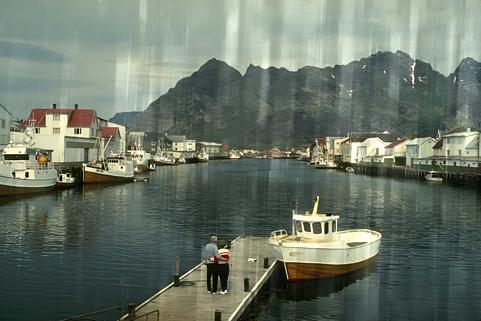}
					\includegraphics[width=.32\linewidth]{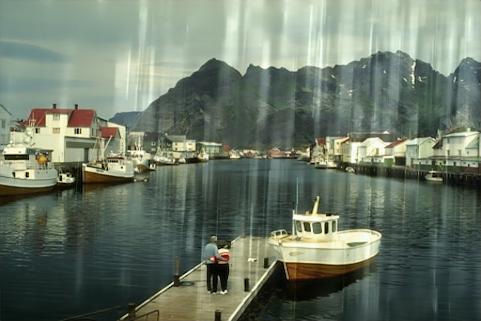}\\ 
					{\large Syn2Real\cite{Yasarla_2020_CVPR} \hskip65pt Syn2Real++\hskip70pt Ground-Truth }\\
					{\large (CVPR'20) \hskip85pt Result \hskip90pt Image }\\
					\includegraphics[width=.32\linewidth]{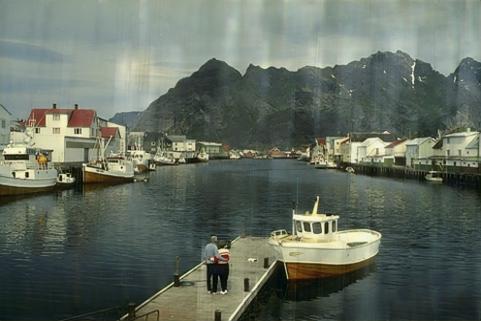}
					\includegraphics[width=.32\linewidth]{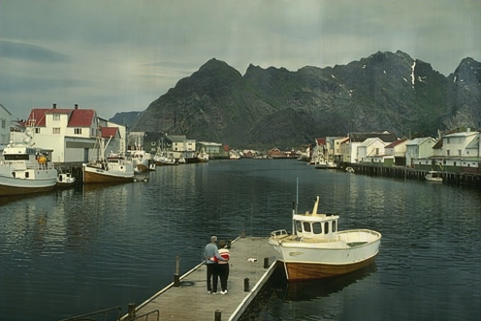}
					\includegraphics[width=.32\linewidth]{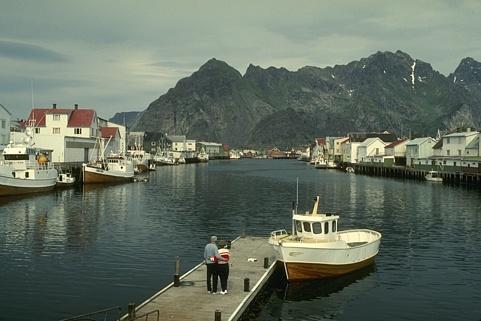}
				\end{center}
			\end{minipage}\\ \vskip10pt 
			\begin{minipage}[c]{0.24\linewidth}
				\begin{center}
					\includegraphics[width=\linewidth]{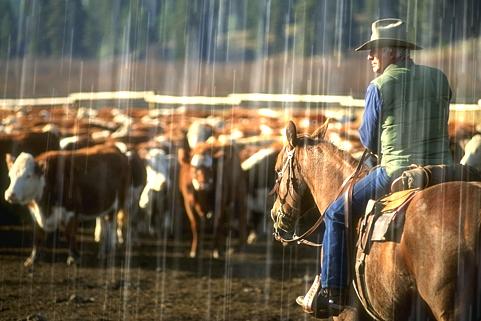}\\
					{\large Rainy Image}
				\end{center}
			\end{minipage}
			\hfill
			\begin{minipage}[c]{0.74\linewidth}
				\begin{center}
					{\large DIDMDN\cite{Authors18} \hskip65pt DDN\cite{Authors17f} \hskip70pt SIRR\cite{wei2019semi} }\\
					{\large (CVPR'18) \hskip65pt (CVPR'17) \hskip70pt (CVPR'19) }\\
					\includegraphics[width=.32\linewidth]{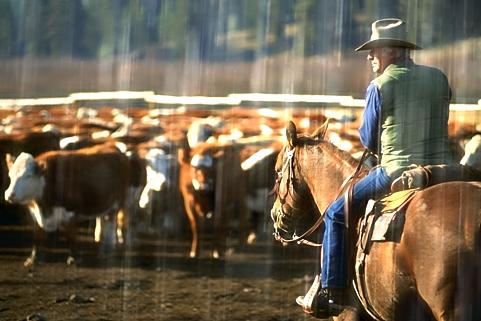}
					\includegraphics[width=.32\linewidth]{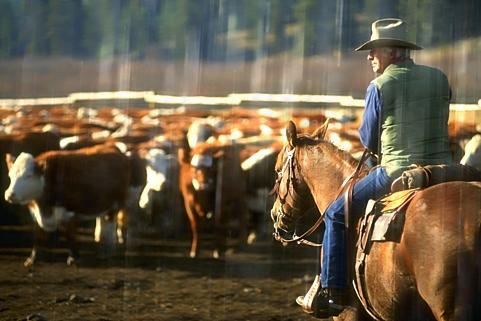}
					\includegraphics[width=.32\linewidth]{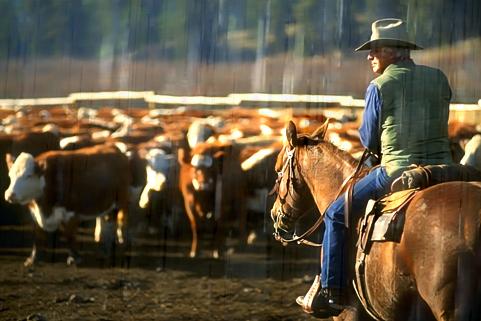}\\ 
					{\large Syn2Real\cite{Yasarla_2020_CVPR} \hskip65pt Syn2Real++\hskip70pt Ground-Truth }\\
					{\large (CVPR'20) \hskip85pt Result \hskip90pt Image }\\
					\includegraphics[width=.32\linewidth]{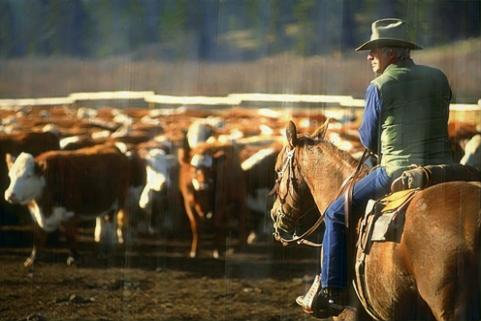}
					\includegraphics[width=.32\linewidth]{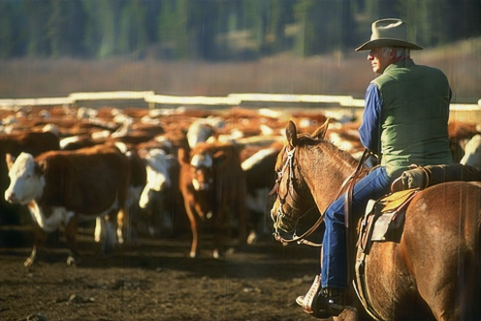}
					\includegraphics[width=.32\linewidth]{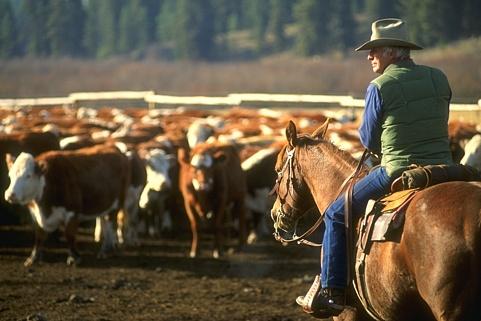}
				\end{center}
			\end{minipage}\\
		\end{center}
		\caption{Qualitative results on DDN-SIRR dataset}
		\label{fig:ddnsirr_synthetic}
	\end{figure*}

		\noindent\textbf{Results on synthetic test set:} The evaluation results on the synthetic test set are shown in Table. \ref{tab:ddnsirr_synthetic}. Similar to \cite{wei2019semi}, we use PSNR as the evaluation metric. We compare the proposed method with several  existing approaches such as DSC \cite{luo2015removing}, LP \cite{li2016rain}, JORDER \cite{yang2017deep}, DDN \cite{Authors17f}, JBO \cite{Authors17c} DID-MDN \cite{Authors18}, and UMRL \cite{yasarla2019uncertainty}. These methods can use only synthetic dataset.  Since the proposed method has the ability to leverage unlabeled real-world data, it is able to achieve significantly better results as compared to the existing approaches. 
		
		Furthermore, we also compare the performance of our method with a recent GMM-based semi-supervised deraining method (SIRR) \cite{wei2019semi}. It can be observed from Table \ref{tab:ddnsirr_synthetic} that the proposed method outperforms SIRR with significant margins. Additionally, we can clearly observe the gains\footnote{The gain is computed by subtracting the performance obtained w/o SSL from the performance obtained using with SSL.} achieved due to the use of additional unlabeled real-world data by both the methods. The proposed method achieves greater gains as compared to SIRR, which indicates that it   has better capacity to leverage unlabeled data. 
		
		Qualitative results on the test set are shown in Fig. \ref{fig:ddnsirr_synthetic}. As can be seen from this figure, \cite{Authors17f,Authors18} under performed in removing the rain streaks. On the other hand SIRR \cite{wei2019semi} which leverages the rain information from real world rainy images by modelling GMM's, was not able remove the rain streaks completely. In our previous approach  (Syn2Real~\cite{Yasarla_2020_CVPR}), we modelled GP in the latent space vector level which assumed same weighted combination for each feature map in $\mathbf{z}$. As shown in the Fig. \ref{fig:ddnsirr_synthetic},   Syn2Real~\cite{Yasarla_2020_CVPR} results are clearer and sharper as compared to the previous methods. However, one may notice the presence of  small rain residues in the sky region  of the first image and near trees in the second image. These issues are overcome in our new approach (Syn2Real++), where we model GP at feature maps level in the latent space. Similarly, we   observe improved performance in terms of quantitative evaluation as well (see Table \ref{tab:ddnsirr_synthetic}).

		\noindent\textbf{Results on real-world test set: } Similar to \cite{wei2019semi,Yasarla_2020_CVPR}, we evaluate the proposed method on the real-world test set from DDN-SIRR. We use no-reference quality metrics NIQE \cite{mittal2012making} and BRISQUE~\cite{mittal2012no} to perform  quantitative  comparison. The results are shown in Table.  \ref{tab:ddnsirr_real}. We compare the performance of our method with SIRR \cite{wei2019semi}, and Syn2Real~\cite{Yasarla_2020_CVPR} which also leverages unlabeled data. It can be observed that the proposed method achieves better performance than SIRR and Syn2Real. Note that lower scores indicate better performance. Furthermore, the proposed method is able to achieve better gains with the use of unlabeled data as compared to SIRR.
	

	\begin{table}[t!]
		\caption{Effect of using unlabeled real-world data in training process on the DDN-SIRR dataset. Evaluation is performed on the \textbf{real-world} test set of DDN-SIRR dataset using no-reference quality metrics (NIQE and BRISQUE). Note that lower scores indicate better performance.}
		\label{tab:ddnsirr_real}
		\centering
		\huge
		\resizebox{1\linewidth}{!}{
			\begin{tabular}{|l|c|ccc|ccc|ccc|}
				\hline
				\multirow{2}{*}{Metrics} & \multirow{2}{*}{\begin{tabular}[c]{@{}c@{}}Input\end{tabular}} & \multicolumn{3}{c|}{SIRR \cite{wei2019semi}} & \multicolumn{3}{c|}{Syn2Real\cite{Yasarla_2020_CVPR}} & \multicolumn{3}{c|}{Syn2Real++} \\ \cline{3-11} 
				&                                                                               & $\mathcal{D_L}$    & $\mathcal{D_L+D_U}$  & Gain  & $\mathcal{D_L}$    & $\mathcal{D_L+D_U}$  & Gain  & $\mathcal{D_L}$    & $\mathcal{D_L+D_U}$  & Gain\\ \hline
				NIQE                                   & 4.671                                                                         & 3.86   & 3.84     & 0.02  & 3.85   & 3.78     & 0.07 & 3.85   & 3.72     & 0.13 \\
				BRISQUE                                & 31.37                                                                         & 26.61  & 25.29    & 1.32  & 25.77  & 22.95    & 2.82 & 25.77  & 22.83    & 2.94 \\ \hline
			\end{tabular}
		} 
	\end{table}

	\begin{figure*}[t!]
		\begin{center}
			\includegraphics[width=.195\linewidth]{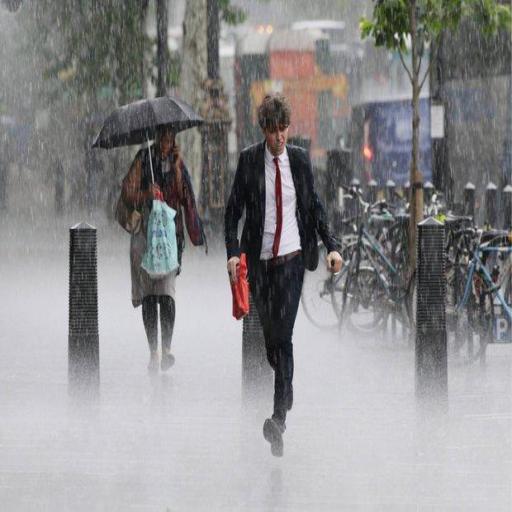}
			\includegraphics[width=.195\linewidth]{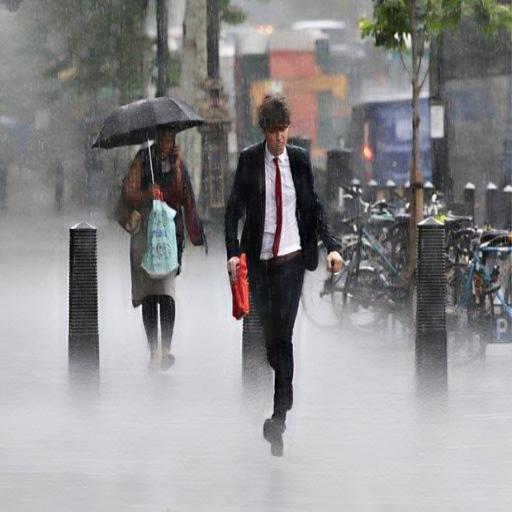}
			\includegraphics[width=.195\linewidth]{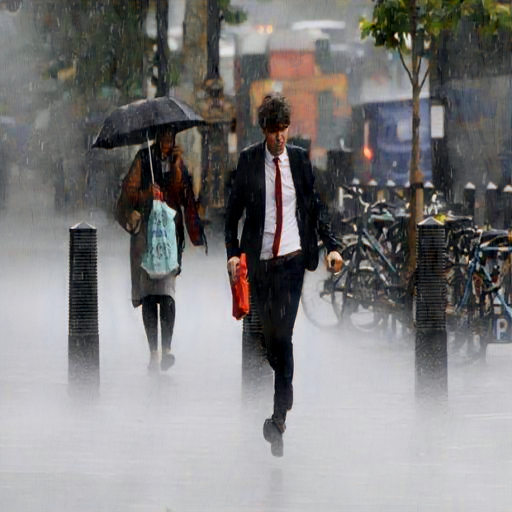}
			\includegraphics[width=.195\linewidth]{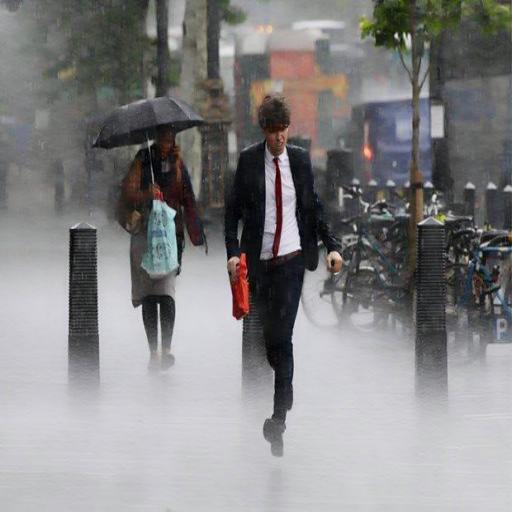}
			\includegraphics[width=.195\linewidth]{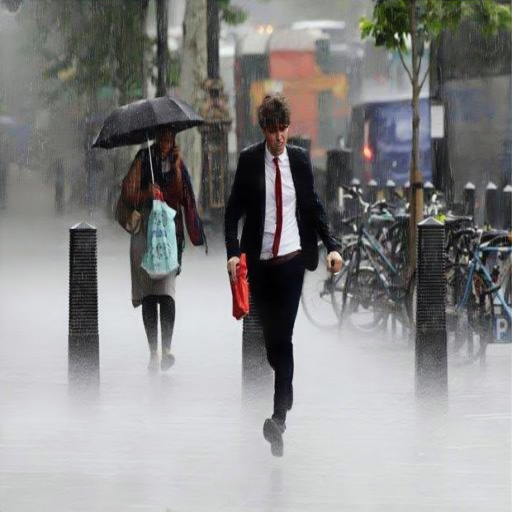}\\ \vskip3pt
			\includegraphics[width=.195\linewidth]{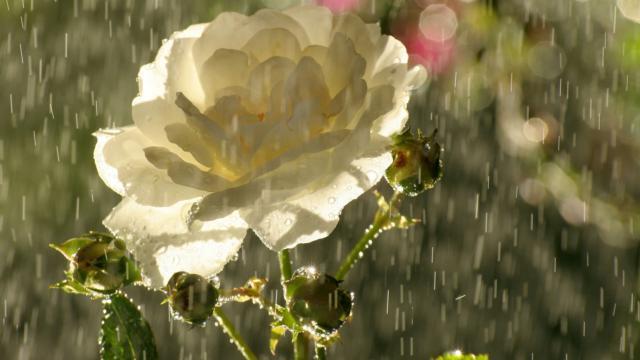}
			\includegraphics[width=.195\linewidth]{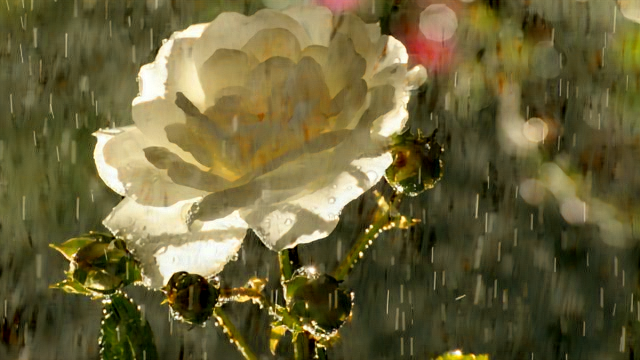}
			\includegraphics[width=.195\linewidth]{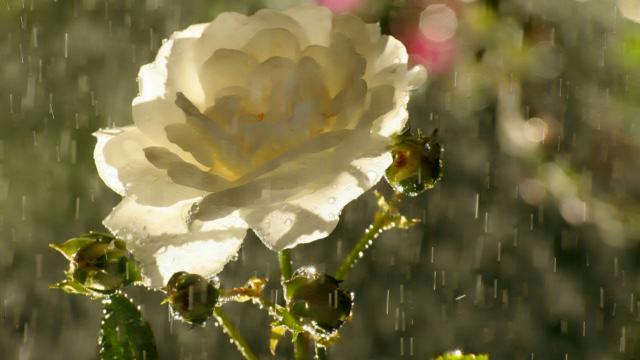}
			\includegraphics[width=.195\linewidth]{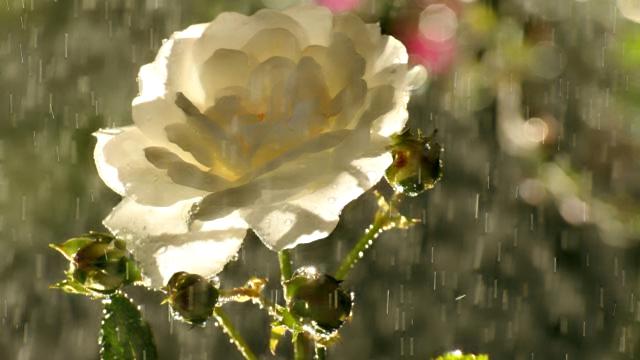}
			\includegraphics[width=.195\linewidth]{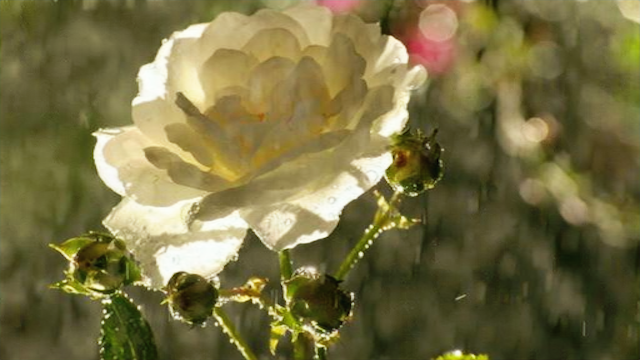}\\ \vskip3pt 
			\includegraphics[width=.195\linewidth]{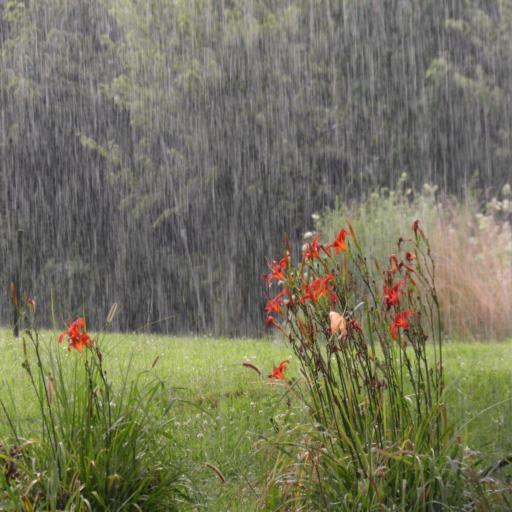}
			\includegraphics[width=.195\linewidth]{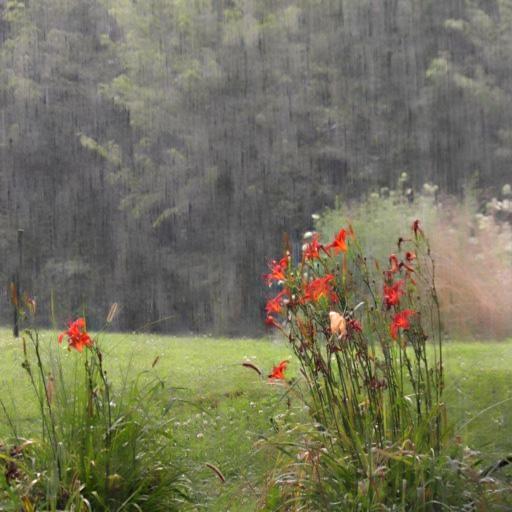}
			\includegraphics[width=.195\linewidth]{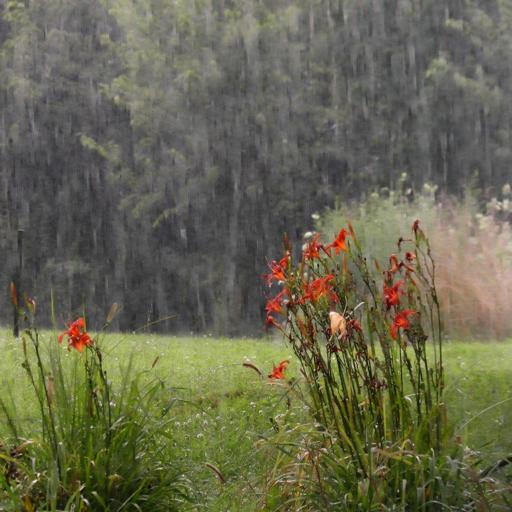}
			\includegraphics[width=.195\linewidth]{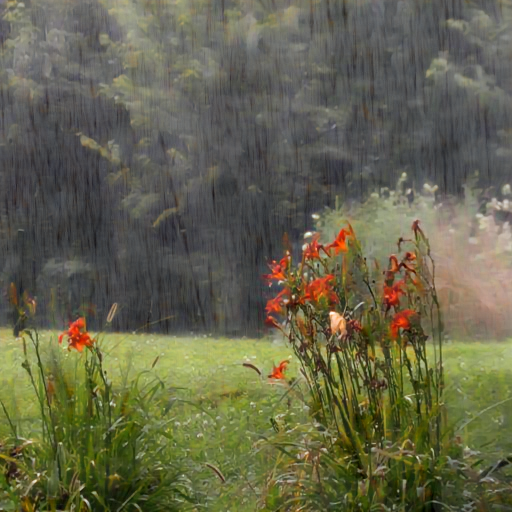}
			\includegraphics[width=.195\linewidth]{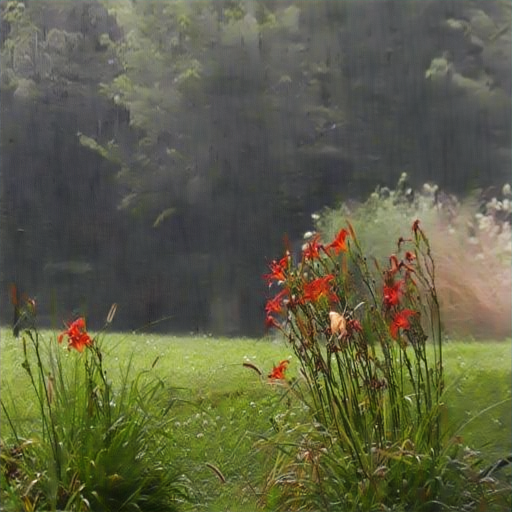}\\ 
			{\large Rainy \hskip45pt DIDMDN\cite{Authors18} \hskip50pt DDN\cite{Authors17f} \hskip50pt SIRR\cite{wei2019semi} \hskip60pt Syn2Real++}\\
			{\large Image \hskip55pt (CVPR'18) \hskip45pt (CVPR'17) \hskip50pt (CVPR'19) \hskip65pt Result}\\
		\end{center}
		\caption{Qualitative results on DDN-SIRR \textbf{real-world} test set. }
		\label{fig:ddnsirr_real}
	\end{figure*}
	
	
	\begin{table*}[t!]
		\caption{Cross domain experiments (PSNR/SSIM): Effect of using unlabelled data $\mathcal{D_U}^{tgt}$ in semi-supervised training process for  leveraging rain information from $\mathcal{D_U}^{tgt}$. Evaluation is performed on the test sets of Rain800, Rain200L and DDN as unlabelled datasets, \textit{i.e} $\mathcal{D_U}^{tgt}$. 	Oracle performance corresponds to the base network trained with $\mathcal{D_{U}}^{tgt}$ in a fully-supervised fashion}
		\label{tab:cross_domain}
		\centering
		\resizebox{.85\linewidth}{!}{
			\begin{tabular}{|l|c|c|c|c|c|c|}
				\hline
				\multirow{2}{*}{Train data } & \multicolumn{3}{c|}{Fully supervised} & \multicolumn{3}{c|}{Semi supervised} \\ \cline{2-7} 
				& $\mathcal{D_U}^{tgt}$ & $\mathcal{D_L}^{src}$ & $\mathcal{D_U}^{tgt}$ & $\mathcal{D_L}^{src}+\mathcal{D_U}^{tgt}$  & $\mathcal{D_L}^{src}+\mathcal{D_U}^{tgt}$ & $\mathcal{D_L}^{src}+\mathcal{D_U}^{tgt}$  \\ \hline
				Test data $\mathcal{D_U}^{tgt}$& PReNet~\cite{ren2019progressive} & Our Base Network & Oracle & SIRR ~\cite{wei2019semi} &  Syn2Real~\cite{Yasarla_2020_CVPR} & Syn2Real++ \\ \cline{1-7} 
				Rain800 & 24.81/0.851 & 21.95/0.716 & 23.75/0.800 & 22.35/0.788 & 22.40/0.748 & 23.24/0.783 \\ \hline
				Rain200L & 32.44/0.941 & 26.10/0.840 & 32.11/0.935 & 25.03/0.842 & 28.24/0.878 & 28.96/0.894 \\ \hline
				DDN & 31.75/0.916 & 26.45/0.831 & 28.98/0.885 & 24.43/0.782 & 27.13/0.845 & 27.47/0.853 \\ \hline
			\end{tabular}
		}
		\vskip3pt
	\end{table*}
	
	Qualitative results on the real world rainy images are shown in Fig. \ref{fig:ddnsirr_real}. As can be seen from this figure, \cite{Authors17f,Authors18} were not able to remove the rain streaks completely due to the the domain gap between synthetic rainy and real-world rainy images, and they are trained on synthetic rainy image datasets. On the other hand SIRR \cite{wei2019semi} which leverages the rain information from real world rainy images by modelling GMM's, was not able remove the rain streaks completely.Using Gaussian processes we are able to leverage the rain information from real-world rainy images and able to produce clean and sharp derained images.

		\subsection{Cross-domain ablation experiments}
		In order to demonstrate that the proposed method can be effectively used for improving cross-dataset/cross-domain deraining performance, we conduct a set of experiments where we train our network on one dataset (source dataset -  $\mathcal{D_L}^{src}$) and transfer it to  another dataset (target dataset - $\mathcal{D_U}^{tgt}$).  We  use the  DIDMDN dataset as the source dataset. We conduct three experiments where we attempt to transfer network trained on the source dataset (DIDMDN) to three target datasets (Rain800, Rain200L and DDN), respectively. The source dataset is considered as labeled dataset  and the target datasets are considered as unlabeled datasets in our framework. The results of these experiments are shown in Table \ref{tab:cross_domain}. 
		

		
		Note, in these experiments we use only rainy images from train split of the corresponding target dataset as unlabeled data $\mathcal{D_U}^{tgt}$ disregarding their corresponding clean images, while training network in unlabeled phase of cross-domain experiments. We test network's performance using the corresponding test set of the unlabeled dataset $\mathcal{D_U}^{tgt}$. Note that PReNet~\cite{ren2019progressive}, our base network and oracle are trained in fully supervised fashion. As shown in  Table~\ref{tab:cross_domain}, we compare our method with PReNet~\cite{ren2019progressive}, SIRR ~\cite{wei2019semi}, and Syn2Real~\cite{Yasarla_2020_CVPR}. In the second column of this table, we show the existing state-of-the-art performance of PReNet~\cite{ren2019progressive} when trained with $\mathcal{D_U}^{tgt}$ in a fully supervised fashion. In the third column, we show the performance of our base network when trained on $\mathcal{D_L}^{src}$ without using GP. Oracle  performance corresponds to our base network trained with $\mathcal{D_{U}}^{tgt}$ in a fully supervised fashion. This clearly shows drop in the performance of the base network when trained on  one dataset and tested on a different dataset. This is due to the domain gap between different rain datasets. In order to improve this cross-dataset performance, we leverage the following   semi-supervised learning frameworks in the training process:
		\begin{itemize}
			\item SIRR~\cite{wei2019semi}: It uses GMM's to leverage the information of rain domain to the other and as shown in the fifth column of Table~\ref{tab:cross_domain}. It is not very effective in leveraging rain information from unlabelled data $\mathcal{D_U}$.
			\item Syn2Real: This corresponds to  our previous approach  where  we use GP to leverage unlabelled rain information. 
			\item Syn2Real++: The proposed extension of Syn2Real where we model  Gaussian process for every feature map in the latent space. 
			\end{itemize}   As it can be observed, the proposed method is  able to perform better than other semi-supervised approaches.

		\begin{figure*}[t!]
			\begin{center}
				\includegraphics[width=.16\linewidth]{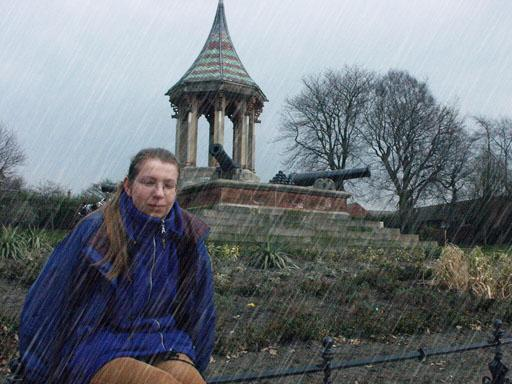}
				\includegraphics[width=.16\linewidth]{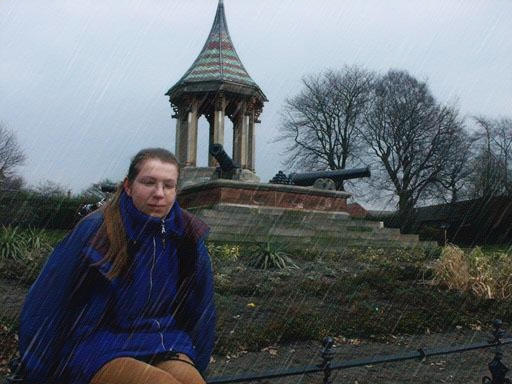}
				\includegraphics[width=.16\linewidth]{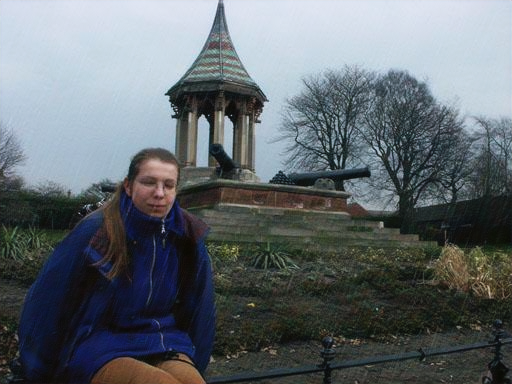}
				\includegraphics[width=.16\linewidth]{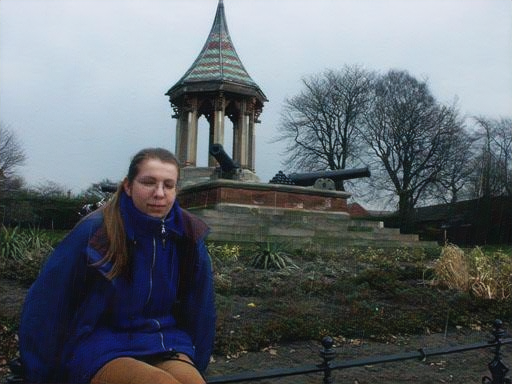}
				\includegraphics[width=.16\linewidth]{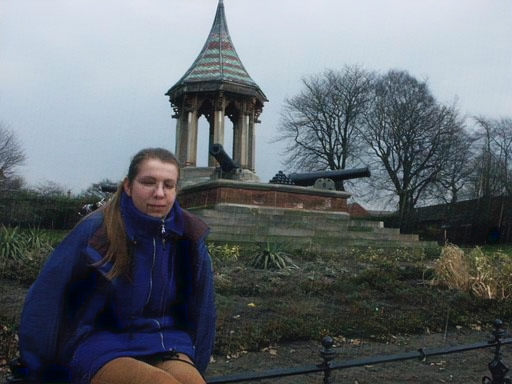}
				\includegraphics[width=.16\linewidth]{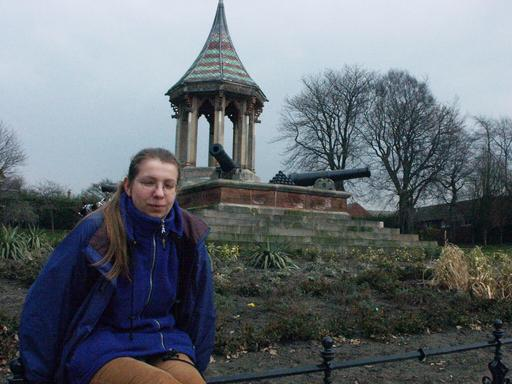}\\ \vskip3pt
				\includegraphics[width=.16\linewidth]{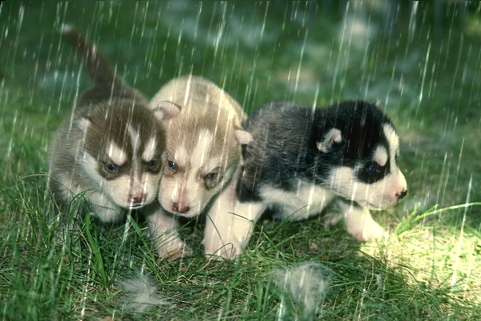}
				\includegraphics[width=.16\linewidth]{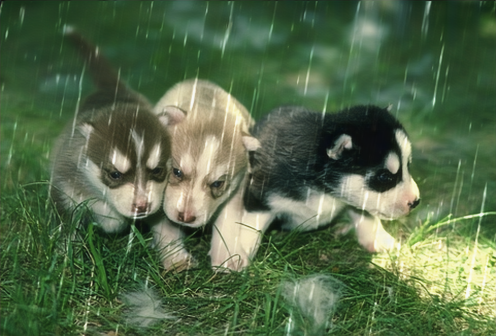}
				\includegraphics[width=.16\linewidth]{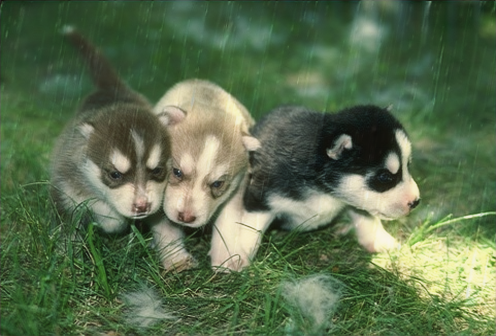}
				\includegraphics[width=.16\linewidth]{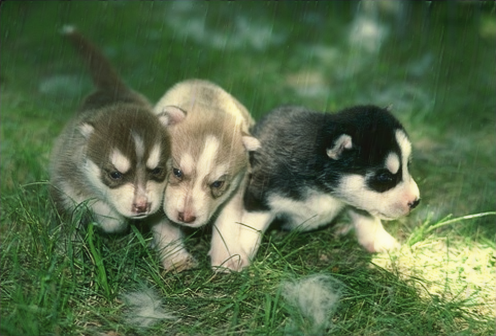}
				\includegraphics[width=.16\linewidth]{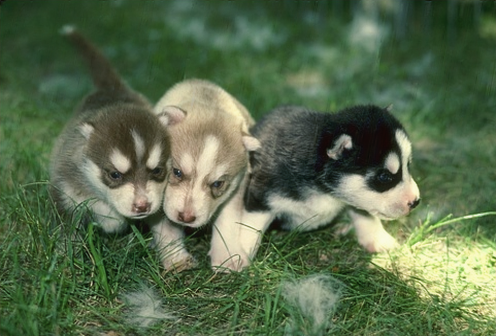}
				\includegraphics[width=.16\linewidth]{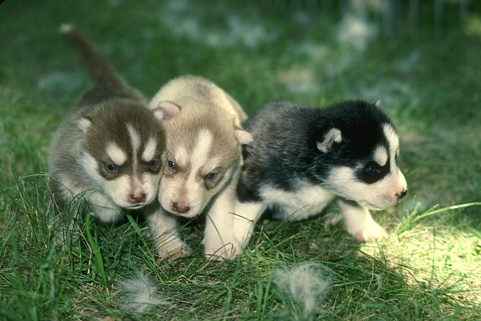}\\ \vskip3pt
				\includegraphics[width=.16\linewidth]{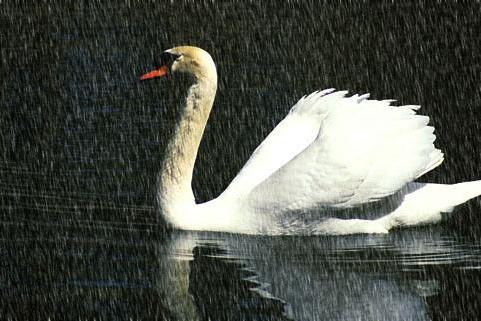}
				\includegraphics[width=.16\linewidth]{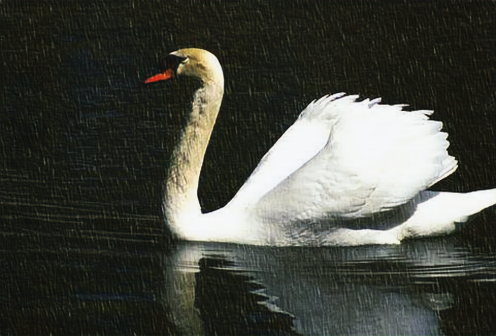}
				\includegraphics[width=.16\linewidth]{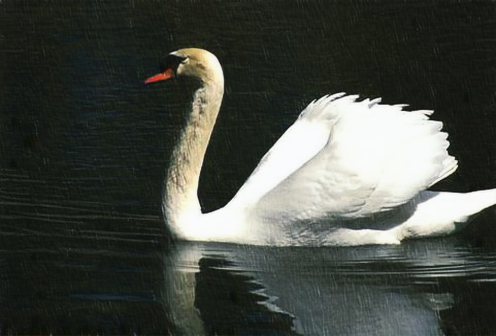}
				\includegraphics[width=.16\linewidth]{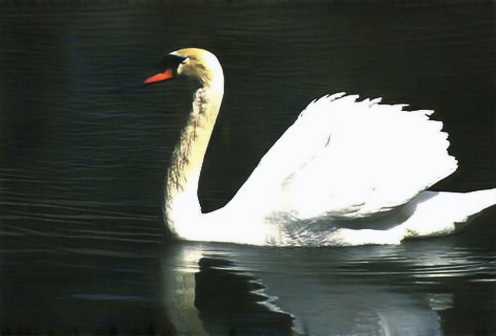}
				\includegraphics[width=.16\linewidth]{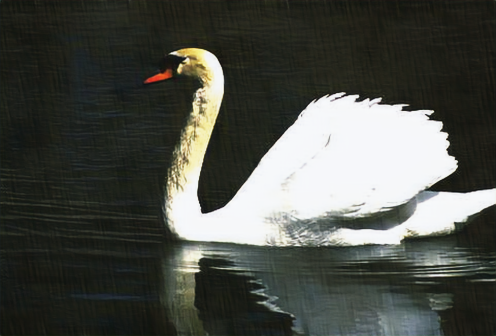}
				\includegraphics[width=.16\linewidth]{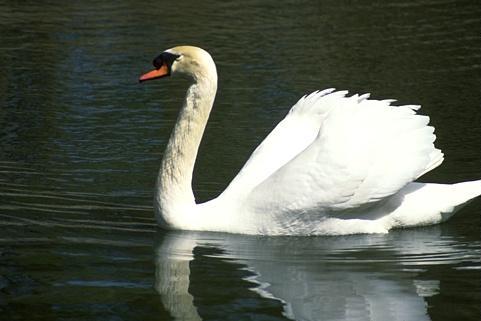}\\
			\end{center}
		    \vskip-8pt
			\hskip10pt Rainy Image  \hskip40pt Without GP\hskip35pt Syn2Real~\cite{Yasarla_2020_CVPR} \hskip30pt Syn2Real++ \hskip35pt Oracle \hskip60pt Clean \\
			\caption{Cross-dataset experiments: Qualitative results on the test sets of Rain800, Rain200L and DDN as unlabelled datasets, \textit{i.e} $\mathcal{D_U}^{tgt}$, where DIDMDN is used as labelled dataset, \textit{i.e} $\mathcal{D_L}^{src}$. Note, oracle is the performance when our base network trained with $\mathcal{D_{U}}^{tgt}$ in fully supervised fashion.}
			\label{fig:crossdomain_abl}
		\end{figure*}
		
		Qualitative results on the test sets of Rain800, Rain200L and DDN as unlabelled datasets, \textit{i.e} $\mathcal{D_U}^{tgt}$  are shown in Fig.~\ref{fig:crossdomain_abl}. As shown in the second column of Fig.~\ref{fig:crossdomain_abl}, the base network when trained on $\mathcal{D_L}^{src}$ without using GP under performed in removing rain streaks. Our previous approach Syn2Real~\cite{Yasarla_2020_CVPR} is able to produce better results but we can still observe few rain streaks in third column of Fig.~\ref{fig:crossdomain_abl}: (i) on the face of woman, and on her coat in the first image, (ii) on the grass in the second image, and (iii) in the water of the third image. The proposed method (Syn2Real++) is able to produce relatively cleaner and sharper derained images, nearly same as the oracle's performance. Note, oracle is the performance when our base network trained with $\mathcal{D_{U}}^{tgt}$ in a fully-supervised fashion.
		\subsection{Hyper-parameters ablation study}
		The goal of these experiments is to analyze the performance of the proposed method for  different set of hyper-parameters. Specifically we provide ablation study for different kernel functions, and different values of nearest neighbors and $\lambda_{unsup}$. In these experiments we use the DIDMDN dataset as labelled, \textit{i.e} $\mathcal{D_L}^{src}$, and  DDN as unlabelled dataset, \textit{i.e} $\mathcal{D_U}^{tgt}$.\\
		
		\noindent\textbf{Kernel Functions: } Kernel function plays an important role in the performance of GPs. In Table~\ref{tab:abl_ker}, we provide the performance of our method using different kernel functions such as Linear kernel (LIN[.]), Squared exponential (SE[.]), and Rational Quadratic (RQ[.]). It can be observed that the performance of the  proposed method is consistent for different kernel functions, with the rational quadratic performing slightly better. However, we use the linear kernel which produces similar results with an additional benefit of being simpler. 
		\begin{table}[h!]
			\caption{Ablation study experiments using different kernel functions Gaussian Process.}
			\label{tab:abl_ker}
			\centering
			\resizebox{1\linewidth}{!}{
				\begin{tabular}{|l|l|c|c|c|}
					\hline
					$\mathcal{D_L}^{src}$ & $\mathcal{D_U}^{tgt}$ & LIN{[}.{]} & SE{[}.{]} & RQ{[}.{]} \\ \hline
					DIDMDN & DDN & 27.47/0.853 & 27.28/0.847 & 27.59/0.851 \\ \hline
			\end{tabular}}
			\vskip3pt
			*PSNR/SSIM metrics are used for comparisons
		\end{table}
		
		\noindent\textbf{Nearest Neighbors $N_n$: } Here we analyze the performance of the network for different values of the nearest neighbours used for obtaining  the pseudo-GT via the GPs. Results corresponding to  this experiment are provided in Table~\ref{tab:abl_near_neigh}. It can be observed that the results are approximately consistent for different numbers of nearest neighbours. $N_n=32$ achieves the best PSNR results, and hence we use it in all our experiments. 
		
		\begin{table}[h!]
			\caption{Ablation study experiments for different number of nearest Neighbors $N_n$.}
			\label{tab:abl_near_neigh}
			\centering
			\resizebox{1\linewidth}{!}{
				\begin{tabular}{|l|l|c|c|c|}
					\hline
					$\mathcal{D_L}^{src}$ & $\mathcal{D_U}^{tgt}$ & $N_n=16$ & $N_n=32$ & $N_n=64$ \\ \hline
					DIDMDN & DDN & 27.17/0.845 & 27.47/0.853 & 27.42/0.856 \\ \hline
			\end{tabular}}
			\vskip3pt
			*PSNR/SSIM metrics are used for comparisons
		\end{table}
		
		\noindent\textbf{Different values of $\lambda_{unsup}$: } Here we analyze the performance of our network for different values of $\lambda_{unsup}$ used for training the network. Results corresponding to  this experiment are provided in Table~\ref{tab:abl_lamb}. It can be observed that $\lambda_{unsup}=1.5 \times 10^{-3}$  achieves the best PSNR/SSIM results, and hence we use it in all our experiments. 
		
		\begin{table}[h!]
			\caption{Ablation study experiments for different values of $\lambda_{unsup}$. }
			\label{tab:abl_lamb}
			\centering
			\resizebox{1\linewidth}{!}{
				\begin{tabular}{|l|l|c|c|c|}
					\hline
					$\mathcal{D_L}^{src}$ & $\mathcal{D_U}^{tgt}$ & $\lambda_{unsup}=1.5 \times 10^{-2}$ & $\lambda_{unsup}=1.5 \times 10^{-3}$ & $\lambda_{unsup}=1.5 \times 10^{-4}$ \\ \hline
					DIDMDN & DDN & 27.09/0.841 & 27.47/0.853 & 27.36/0.849 \\ \hline
			\end{tabular}}
			\vskip3pt
			*PSNR/SSIM metrics are used for comparisons
		\end{table}

	\section{Conclusion}
	We presented a GP-based SSL framework to  leverage unlabeled data during training for the image deraining task. We use supervised loss functions such as $l_1$ and the perceptual loss to train on the labeled data. For the unlabeled data, we estimate the pseudo-GT at the latent space  by jointly modeling the labeled and unlabeled latent space vectors using the GP. The pseudo-GT is then used to supervise for the unlabeled samples. Through extensive experiments on several datasets such as DDN, Rain800, Rain200L and DDN-SIRR, we demonstrate that the proposed method is able to achieve better generalization by leveraging unlabeled data.

	\ifCLASSOPTIONcaptionsoff
	\newpage
	\fi

	
	
	%

	\bibliography{egbib}
	\bibliographystyle{IEEEtran}

\end{document}